\documentclass[12pt]{article}

\usepackage{graphicx}%
\usepackage{multirow}%
\usepackage{amsmath,amssymb,amsfonts}%
\usepackage{amsthm}%
\usepackage{mathrsfs}%
\usepackage[title]{appendix}%
\usepackage{xcolor}%
\usepackage{textcomp}%
\usepackage{manyfoot}%
\usepackage{booktabs}%
\usepackage{algorithm}%
\usepackage{algorithmicx}%
\usepackage{algpseudocode}%
\usepackage{listings}%
\usepackage{mathtools}
\usepackage{rotating}
\usepackage{lscape}
\usepackage{amsfonts}
\usepackage{pifont}
\usepackage{csquotes}
\usepackage{array} 
\usepackage{pifont} 
\usepackage{hyperref} 
\usepackage[numbers]{natbib}
\usepackage{colortbl}
\usepackage{array}
\usepackage{rotating}
\usepackage{subcaption}
\usepackage{float} 
\usepackage{adjustbox,lipsum}
\usepackage{parskip}
\usepackage{placeins} 
\usepackage{lmodern}     

\title{Imputation of Longitudinal Data Using GANs: Challenges and Implications for Classification}

\author{
\makebox[\textwidth]{\parbox{0.85\textwidth}{\centering
\textbf{1. Sharon Torao Pingi}\textsuperscript{*}\\
\textbf{2. Md Abul Bashar}, \textbf{3. Richi Nayak} \\[0.5em]
School of Computer Science \& Centre for Data Science \\
Queensland University of Technology \\[0.5em]
\texttt{sharon.torao@hdr.qut.edu.au}, \\
\texttt{m1.bashar@qut.edu.au}, \texttt{r.nayak@qut.edu.au}
}}
}

\date{}  

\begin{document}
\maketitle


\begin{abstract}
Longitudinal data is commonly utilised across various domains, such as health, biomedical, education and survey studies. This ubiquity has led to a rise in statistical, machine and deep learning-based methods for Longitudinal Data Classification (LDC). However, the intricate nature of the data, characterised by its multi-dimensionality, causes instance-level heterogeneity and temporal correlations that add to the complexity of longitudinal data analysis. Additionally, LDC accuracy is often hampered by the pervasiveness of missing values in longitudinal data. Despite ongoing research that draw on the generative power and utility of Generative Adversarial Networks (GANs) to address the missing data problem, critical considerations include statistical assumptions surrounding longitudinal data and missingness within it, as well as other data-level challenges like class imbalance and mixed data types that impact longitudinal data imputation (LDI) and the subsequent LDC process in GANs. This paper provides a comprehensive overview of how GANs have been applied in LDI, with a focus whether GANS have adequately addressed fundamental assumptions about the data from a LDC perspective. We propose a categorisation of main approaches to GAN-based LDI, highlight strengths and limitations of methods, identify key research trends, and provide promising future directions. Our findings indicate that while GANs show great potential for LDI to improve usability and quality of longitudinal data for tasks like LDC, there is need for more versatile approaches that can handle the wider spectrum of challenges presented by longitudinal data with missing values. By synthesising current knowledge and identifying critical research gaps, this survey aims to guide future research efforts in developing more effective GAN-based solutions to address LDC challenges.
\end{abstract}



\section{Introduction} 
\label{introduction}
Longitudinal data is prevalent in many fields, including critical sectors like health and education. For example, it supports tracking and detecting Alzheimer's disease in patients within electronic health records (EHR) or monitoring student progress in education data for early signs of school dropout. These time-sensitive applications follow the same subjects or instances over time to monitor changes in certain variables for tasks like forecasting or classification.

Longitudinal data is inherently multi-dimensional, consisting of a non-temporal component that includes static features (e.g. gender or race, not expected to change over time), and a temporal component that often resembles multi-variate time series (e.g. blood pressure, heart rate). However, the temporal component can also take the form of images like MRIs, sequences like RNA, or prescription records, measured repeatedly for the same instances (e.g. patients) over time. This survey focuses exclusively on tabular longitudinal data, which is most widespread, where the time series component can comprise a mix of discrete and continuous features, depending on whether the time dimension is unbounded or bounded \cite{Brophy2023gantssurvey}. 

Beyond their multidimensional nature, longitudinal data analysis tasks like LDC face additional challenges, including statistical assumptions about the data and endemic data-level issues. Key data-level challenges include missing values, class imbalance, mixed data types, and small dataset sizes. While some of these issues can be mitigated by resampling minority classes or collecting more data, missing values poses the most significant problem. Missing data in longitudinal datasets can arise from random processes like data entry errors, however is often caused by irregular sampling and study drop outs that present in the time series component. Local and global correlations exist within the temporal component and are affected by missing data patterns.  

In this survey paper, we focus on the missing data problem, particularly in the temporal component for several reasons: 1) it has the most significant impact on LDC performance compared to other data-level challenges \cite{jahangiri2023imputation}, due to disrupted temporal correlations \cite{sun2020irregular}, 2) improper handling of missing values can interfere with underlying temporal patterns and relationships \cite{hassankhani2023imputation}, leading to sub-optimal LDC accuracy \cite{Palanivinayagam2023missingvalues}, 3) sometimes informative nature of missing data patterns \cite{little2019missing}, 4) no universally optimal imputation technique exists, given each longitudinal dataset's unique temporal patterns \cite{pingi2022logan,Ribeiro2021imputation}, 5) the complexity required for contextual imputation given instance or group level heterogeneity, and 6) effectively addressing missing data significantly enhances data quality and usability, thereby improving the generalisation and robustness of LDC models \cite{jahangiri2023imputation}. We also consider the assumptions surrounding missing data in longitudinal datasets \cite{little2013missing} and their implications on longitudinal data imputation (LDI). 

In this paper, we use the term \enquote{imputation} more broadly than its conventional meaning of filling in missing data with estimated values. Here, it refers to any method used to address the missing data problem. We specifically refer to missing temporal data as Incomplete Time Series (ITS) and the imputed or reconstructed time series generated by GAN-based LDI methods (or other means) as Complete Time Series (CTS). We focus on the time series aspect because most works in this field address missingness primarily in this component, often ignoring or under-utilising the static component in the modelling process.

LDC research has traditionally focused on discriminative modelling approaches. To address missingness in longitudinal data, common methods include statistical techniques like multiple imputation \cite{de2017statldcmiss}, maximum likelihood \cite{pennoni2022mlelongmiss}, and mixed effects models \cite{koelmel2017melongmiss}. These methods often struggle to scale to large datasets and require extensive feature engineering. In contrast, deep learning models have proven more powerful in handling large, complex data structures, learning inherent relationships and feature correlations \cite{cao2018brits,xu2021rnnmiss}. Despite their strengths, discriminative models make assumptions of sample homogeneity and balanced class distributions  \cite{bennett2022imbalance}, and they are not inherently designed to handle missing data directly \cite{ng2002classification}. While some discriminative methods can handle longitudinal data of uneven lengths \cite{Qin2023gatedrnn,rubanova2019odernn}, they still focus on finding decision boundaries between classes, and like other deep learning methods, do not focus on learning the underlying data distribution, which is crucial for generating more context-aware estimates for missing data. 

Deep Generative Models (DGMs) \cite{salakhutdinov2015generative} such as deep auto-regressive models \cite{salinas2020deepar}, variational autoencoders (VAE) \cite{kingma2013vae}, and especially Generative Adversarial Networks (GANs) \cite{goodfellow2014gan} have been effectively used to address some of the missing data challenges mentioned. DGMs like the VAE maximises the likelihood of observing the training data by estimating a tractable density function and optimising a variational lower bound on the likelihood. In contrast, GANs implicitly learn model parameters through an adversarial training process, which, when stabilised, produces better samples \cite{bond2021surveydgm}. 

In this survey paper, we focus on GAN-based methods for LDI due to their 1) widespread use compared to other DGMs for imputing longitudinal and time series data, 2) ability to learn complex data distribution and generate context-dependent estimates for missing values, 3) flexibility to adapt to specific tasks like LDI, for instance by incorporating missing data information during distribution learning \cite{ipsen2022missing}, 4) adaptability across multiple tasks, e.g. with joint auxiliary like LDC \cite{ma2020ajrnn} and 5) capacity to converge to optimal models based on the data and mode parameters ($P(X,\theta)$). These advantages give GANs a clear edge in handling missing values for LDC over other methods. 

For LDC, GANs have been used to address missing values either during the preprocessing stage, with classification as a downstream task, or as an end-to-end approach where a classifier is integrated into the GAN model. In these joint optimisation strategies, the GAN model learns to model a conditional data distribution that assists primary objectives like imputation \cite{ma2020ajrnn,sun2020irregular}. However, some GAN-based approaches for longitudinal data generation bypass the missing data issue by discarding instances with missing values \cite{Bernardini2023ccgan}, which leads to data loss and ignores correlations between observations and missing patterns. Under certain conditions (see Section \ref{sec:challengs_mvs}), missing data patterns are informative, GAN-based LDI methods should consider the missingness mechanism \cite{little2013missing}, as well as instance heterogeneity, temporal correlations, and other data-level challenges. This holistic approach is essential to avoid sub-optimal analysis and to maximise classifier performance \cite{hamel2012lng,marti2020lng}. 

Table \ref{tab:surveys} summarises the related survey papers. A review of  these works highlights various applications of GANs for imputation across different data types, including general imputation \cite{Kim2020ganmvsurvey,Shahbazian2023ganmvsurvey}, specific types of longitudinal data such as EHR \cite{Psychogyios2023ehrmvsurvey, Ghosheh2024surveygansehr}, and time series data \cite{Brophy2023gantssurvey, Festag2022ganbiomedtssurvey}. Additionally, some surveys focus on specific types of GANs, like the Generative Adversarial Imputation GAN (GAIN) \cite{yoon2018gain} for tabular data imputation \cite{Zhang2023gainsurvey}. Other reviews cover general machine learning or deep learning-based imputation methods for EHR \cite{Kazijevs2023mvtsehrsurvey}, time series data \cite{Sun2023dltsmvsurvey,Liu2023dlhealthmvsurvey}, and other domains and applications \cite{Cascarano2023dllongbiomed}. 

While works exist on 1) GANs for imputation in general, 2) imputation of only specific types of longitudinal data, like EHRs, and 3) general application of GANs for longitudinal datasets, to the best of our knowledge, this is the first paper that specifically surveys GANs for imputing missing longitudinal data, focusing on the unique characteristics of the datasets, the assumptions surrounding missing longitudinal data, and other data-level challenges that hinder LDC performance. 
More specifically, our contributions are as follows:
\begin{enumerate}
    \item We provide a topology of data-level challenges for LDC from a GAN perspective.
    \item We synthesise existing research to highlight key themes and challenges addressed in LDI using GANs.
    \item We propose a classification of GAN-based approaches for imputation of longitudinal data. 
    \item  We present a meta-analysis of existing works to identify past, current research trends and future research directions for GANs for LDI and LDC.
    \item We summarise our findings and identify gaps in the literature to guide future research. 
\end{enumerate}

\begin{table}[!ht]
\footnotesize 
\centering
\begin{minipage}{\columnwidth} 
\caption{Existing survey works on GANs for longitudinal data imputation (LDI)}
\label{tab:surveys}
\begin{tabular}{|l|c|l|l|l|}
\hline
\textbf{Paper} &
  \textbf{Year} &
  \textbf{\begin{tabular}[c]{@{}l@{}}GANs + \\      Imputation + \\      longitudinal  \\data\end{tabular}} &
  \textbf{Domain} &
  \textbf{Summary of   contributions} \\ \hline
\begin{tabular}[c]{@{}l@{}}\cite{Kim2020ganmvsurvey}\end{tabular} &
  2020 &
  \ding{51},   \ding{51}, \ding{55} &
  Any &
  \begin{tabular}[c]{@{}l@{}}Brief   coverage of GANs for missing data \\      imputation, mostly for image data. \\      Brief on a GAN for EHR.\end{tabular} \\ \hline
\begin{tabular}[c]{@{}l@{}}\cite{Xie2022ehrsurvey}\end{tabular} &
  2020 &
  \ding{55},   \ding{55}, \ding{51} &
  EHR &
  \begin{tabular}[c]{@{}l@{}}Deep   learning for temporal data \\      representation in EHR. Focus on \\      irregularity, heterogeneity, sparsity. \\      None on GANs, brief on imputation.\end{tabular} \\ \hline
\begin{tabular}[c]{@{}l@{}}\cite{Ghosheh2022surveygansehr}\end{tabular} &
  2022 &
  \ding{51},   \ding{55}, \ding{51} &
  EHR &
  \begin{tabular}[c]{@{}l@{}}GANs   for EHR, applications in data \\      synthesising, and other applications. \\      Brief on imputation.\end{tabular} \\ \hline
\begin{tabular}[c]{@{}l@{}}\cite{Festag2022ganbiomedtssurvey}\end{tabular} &
  2022 &
  \ding{51},   \ding{51}, \ding{55} &
  \begin{tabular}[c]{@{}l@{}}Biomedical   \\      time series\end{tabular} &
  \begin{tabular}[c]{@{}l@{}}GANs   for missing data imputation \\      \& forecasting of biomedical time series. \\      Brief on EHR.\end{tabular} \\ \hline
\begin{tabular}[c]{@{}l@{}}\cite{Shahbazian2023ganmvsurvey}\end{tabular} &
  2023 &
  \ding{51},   \ding{51}, \ding{55} &
  Any &
  \begin{tabular}[c]{@{}l@{}}GANs   for imputation. Focus on missing \\      data type, missing  fraction   \&  imputation \\      algorithm. Brief on longitudinal data. \\      Incl. experiments.\end{tabular} \\ \hline
\begin{tabular}[c]{@{}l@{}}\cite{Brophy2023gantssurvey}\end{tabular} &
  2023 &
  \ding{51},   \ding{55}, \ding{55} &
  Time   series &
  \begin{tabular}[c]{@{}l@{}}Time   series-related applications of \\      GANs. Proposes GAN variants: continuous \\      and discrete, based on time series type. \\      Brief on imputation.\end{tabular} \\ \hline
\begin{tabular}[c]{@{}l@{}}\cite{Sun2023dltsmvsurvey}\end{tabular} &
  2023 &
  \ding{55},   \ding{51}, \ding{55} &
  \begin{tabular}[c]{@{}l@{}}Medical   \\      time series\end{tabular} &
  \begin{tabular}[c]{@{}l@{}}Deep   learning methods for irregular \\      medical time series. Focus on intra / inter-\\      series relationships \& local / global \\      correlations. Brief on GANs. Inc. experiments.\end{tabular} \\ \hline
\begin{tabular}[c]{@{}l@{}}\cite{Liu2023dlhealthmvsurvey}\end{tabular} &
  2023 &
  \ding{55},   \ding{51}, \ding{55} &
  \begin{tabular}[c]{@{}l@{}}Healthcare,   \\      time series\end{tabular} &
  \begin{tabular}[c]{@{}l@{}}Deep   learning for missing data in \\      healthcare data. Includes images, genome, \\      signals and multi-modal data.  Brief on GANs\\       and longitudinal data.\end{tabular} \\ \hline
\begin{tabular}[c]{@{}l@{}}\cite{Cascarano2023dllongbiomed}\end{tabular} &
  2023 &
  \ding{55},   \ding{55}, \ding{51} &
  \begin{tabular}[c]{@{}l@{}}Longitudinal   \\      biomedical\end{tabular} &
  \begin{tabular}[c]{@{}l@{}}Machine   and deep learning methods and \\      applications for longitudinal data biomedical \\      data. Brief on GANs.\end{tabular} \\ \hline
\begin{tabular}[c]{@{}l@{}}\cite{Zhang2023gainsurvey}\end{tabular} &
  2023 &
  \ding{55},   \ding{51}, \ding{55} &
  Any &
  \begin{tabular}[c]{@{}l@{}}GAIN-based \cite{yoon2018gain} only. Other GANs \& deep \\       learning methods discussed only regarding \\      GAIN. Brief on longitudinal data \& imputation.\end{tabular} \\ \hline
\begin{tabular}[c]{@{}l@{}}\cite{Kazijevs2023mvtsehrsurvey}\end{tabular} &
  2023 &
  \ding{55},   \ding{55}, \ding{51} &
  EHR &
  \begin{tabular}[c]{@{}l@{}}Benchmarking   of deep imputation methods \\      for time series health data. GAIN-based only. \\ Includes experiments.\end{tabular} \\ \hline
Ours &
  2024 &
  \ding{51},   \ding{51}, \ding{51} &
  Longitudinal &
  \begin{tabular}[c]{@{}l@{}}GANs  addressing missing values for improved \\       longitudinal data classification. Topology of \\data-level challenges. Categorise imputation \\approaches. Focus on fundamental assumptions \\ \& challenges in longitudinal data.\end{tabular} \\ \hline
\end{tabular}%
\end{minipage}
\end{table}

The rest of this paper is organised as follows. Section \ref{sec:background} presents background on longitudinal datasets and LDC, common GAN architectures used for LDI and LDC, and the two main approaches to LDC in GANs. Section \ref{sec:survey_methodology} outlines the survey methodology. 
Section \ref{sec:LDC_GAN_challenges} discusses the challenges LDC poses for GANs on longitudinal data, as well as challenges specific to GAN models. Section \ref{sec:ldi_gan_approaches} reviews approaches and state-of-the-art methods for LDI using GANs. We then present relevant findings, research implications and future directions in Section \ref{sec:discuss_future}, followed by the conclusion. 

\section{Background: LDC with missing values and GAN} 
\label{sec:background}
In this section, we provide essential context on the inherent features of longitudinal datasets and the resulting challenges for LDC, particularly from a missing data standpoint. Formal definitions and a table of symbols and acronyms (Table \ref{tab:symbols}) used throughout the paper are provided. We also introduce seminal GANs most commonly utilised for time series or longitudinal data imputation. Finally, we examine the general challenges GANs face, and their address in GANs for LDI. 

\begin{table}[!ht]
    \centering
    \small
    \caption{Table of Symbols and Acronyms} 
    \begin{tabular}{|c|l||c|l|} \hline 
        \multicolumn{1}{|c|}{\textbf{\begin{tabular}[c]{@{}c@{}}Sym /   \\      Acron\end{tabular}}} & \multicolumn{1}{c||}{\textbf{Meaning}} & \multicolumn{1}{c|}{\textbf{\begin{tabular}[c]{@{}c@{}}Sym /   \\      Acron\end{tabular}}} & \multicolumn{1}{c|}{\textbf{Meaning}} \\ \hline
        $\Delta$ & Temporal Decay & GRU & Gated Recurrent Unit \\ \hline
        AE & Autoencoder & GRU-D & GRU with trainable decays \\ \hline
        $\mathbb{C}$ & Critic in a Wasserstein GAN & GRUI & GRU for imputation \\ \hline
        $D$ & Discriminator & ITS & Incomplete Time Series \\ \hline
        $G$ & Generator & LDC & Longitudinal Data Classification \\ \hline
        $M$ & Missing data mask & LDI & Longitudinal Data Imputation \\ \hline
        Bi- & Bi-directional (e.g. BiRNN) & MAR & Missing At Random \\ \hline
        BiGAN & Bi-directional GAN & MCAR & Missing Completely At Random \\ \hline
        cGAN & Conditional GAN & MNAR & Missing Not At Random \\ \hline
        CTS & Complete Time Series & MTS & Multi-variate Time Series \\ \hline
        DAE & Denoising AE & RNN & Recurrent Neural Networks \\ \hline
        EHR & Electronic Health Records & VAE & Variational AE \\ \hline
        FCN & Fully Connected Network & WGAN & Wasserstein GAN \\ \hline
        GAN & Generative Adversarial Network & Z & Noise vector \\ \hline
    \end{tabular}
    \label{tab:symbols}
\end{table}

\subsection{Longitudinal data and LDC}
\label{sec:bg_ld_ldc}
\begin{figure}[!ht]
    \centering
    \includegraphics[width=0.8\linewidth]{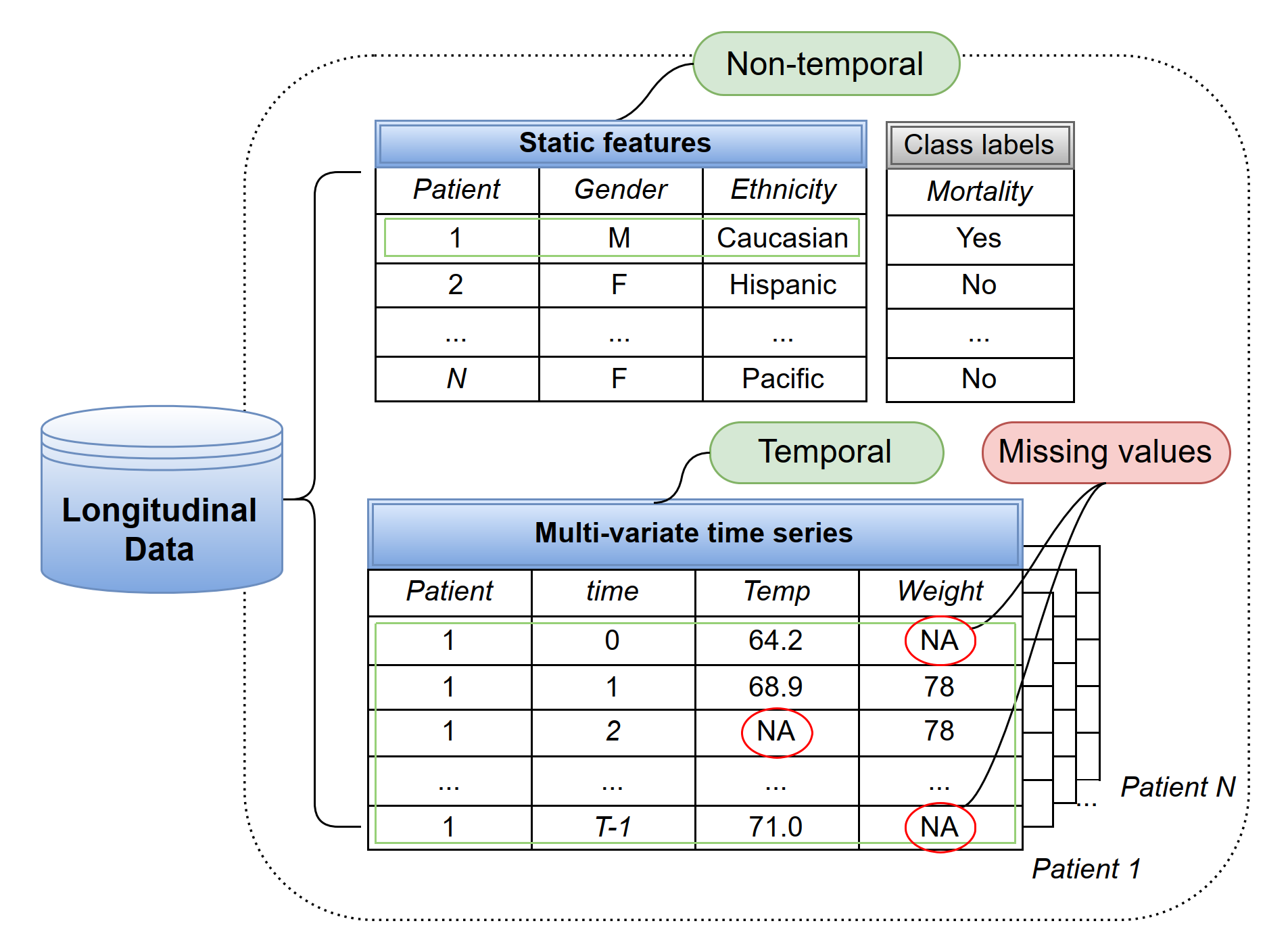}
    \caption{Simplistic example of longitudinal data with static and temporal data components, with illustration of missing values.}
    \label{fig:longdata}
\end{figure}

The defining characteristic of longitudinal data is the repeated observation of subjects or entities over time, referred to as \textit{instances} throughout this paper. In tabular longitudinal datasets (shown in Figure \ref{fig:longdata}), multivariate time series make up its temporal dimension, and features measured at baseline are static and not expected to change with time. Temporal observation times, or \textit{time steps}, can vary between instances. This paper considers both continuous and discrete time series. Continuous time series arise from processes observed over a continuous time domain, such as EEG signals. In contrast, discrete time series are sampled at specific, finite time points, resulting in a countable time domain, as seen in blood pressure measurements. The static and temporal features may consist of mixed data types.

Instances are assumed to be i.i.d., meaning that static and temporal features of individual instances are independent and identically distributed across the population, allowing for population-level inferences based on instances observed. However, the static features(observed) and \textit{random effects} (unobserved) contribute to \textit{instance-level heterogeneity} within temporal features from the same instance. Thus, individual instance's multivariate time series are non-i.i.d. and exhibit intra-instance correlations \cite{fitzmaurice2012long}. Also, within each time series, there is \enquote{auto-correlation}, where past data correlate with current and future observations.

In many research or stakeholder contexts, instances must be grouped into categories, such as individuals diagnosed with dementia, those transitioning into dementia and those without dementia \cite{marcus2010oasis}. When instances are unevenly distributed across categories, the dataset is considered \textit{imbalanced}. The goal of classification tasks like LDC is to train a model that can accurately assign unseen instances to these groups, which requires a labelled dataset. The left of Figure \ref{fig:framework} shows a simplistic example of a longitudinal dataset. There are two static features (Gender \& Race), and two temporal features (Temperature \& Weight) with missing values ($NA$), and mortality as the class outcome. 

\subsubsection{LDC-related Definitions} \label{sec:definitions}
\label{sec:bg_ldc_defns}
LDC aims to minimise the misclassification rate of the class label $y$ based on the posterior distribution $p(y|X)$, where $X$ represents the observed data. Whilst discriminative models approach this directly by defining the decision boundary between classes, generative models like GANs first estimate the joint distribution $p(X,y)$ before determining the most likely $y$. Despite their different methodologies, the classification objective remains the same in both approaches. \\


\noindent \textbf{Definition 1:} \textit{Longitudinal Data} \\
Consider a collection $\mathcal{S}$ of longitudinal samples, where $i \in \mathcal{S}$ is an instance in the collection. Each instance consists of $Q$ static variables $x_i^q$, i.e. $X_i^Q = \{x_i^q\}_{q=1}^Q$ and $V$ temporal variables $x_i^v$, i.e. $X_i^V = \{x_i^v\}_{v=1}^V$. 
Each temporal variable is observed at time points $t$ over a period $T$. The length of $T$ can vary across instances, i.e. for a given instance, the temporal dimension for some variables may be less than $V$.  If the temporal component is modelled as a multivariate time series (MTS) with the class labels $y = \{1,..K\}$ for $K$ number of classes, then an instance's complete record is given by $X_i$ = \{$X_i^Q, X^{V}_{i,1:T}, y_i$\}, or \{$X^Q, X^{V}_{1:T}, y$\} if we ignore $i$ index for notation simplicity. \\

\noindent \textbf{Definition 2:} \textit{Irregular Sampling} \\
Occurs when temporal variables $v$ are observed at irregular time intervals for each instance. When regularly sampled, the MTS has dimensions $[V \times T]$ for all instances, resulting in temporal alignment across the dataset. As noted by \cite{shukla2020surveyirregts}, irregularly sampled time series can be represented as: 1) a set of time-value paired observations where time order is unimportant, 2) a set of univariate time series with uneven lengths where time order is important, or 3) a discretised vector that enforces regular sampling, resulting in missing values (NULL) at time points $t$ where $x^v$ is unobserved. Thus, an irregularly sampled time series $x_t^v$ could be represented as:
\begin{equation}
  x_t^v =
    \begin{cases}
      x_{t}^v, & \text{if } x_t^v \text{ is observed }\\
      \text{NULL}, & \text{otherwise}
    \end{cases} 
    \label{eqn:vectorrep}
\end{equation}
A binary \textit{missingness data mask} vector $M$ of size $[V \times T]$ is commonly used to represent the presence or absence of data. Each temporal variable $X^v$ can then be represented as a pair $(m_t^v,x_t^v)$ for each time point $t$, where the missingness at time $t$ for $x^v_t$ is given as follows:
\begin{equation}
  m_t^v =
    \begin{cases}
      1, & \text{if } x_t^v \text{ is observed }\\
      0, & \text{otherwise}
    \end{cases}  
    \label{eqn:missmask}
\end{equation}

Temporal decay mechanisms \cite{che2017grud} that capture \textit{time lags} between observations $x_t^v$ are given as $\delta_t^v \in \Delta$. If sampling is regular, time is treated as an additional dimension, and $\Delta$ has a size of $[V \times T]$. In this discretised representation, $T$ is not explicitly recorded. Otherwise, timestamps $s_t$ are recorded alongside each observation. In such cases, the time lags $\delta_t^v$ are calculated as: 
\begin{equation}
  \delta_{t}^{v}=
    \begin{cases}
        s_{t}-s_{t-1}+{\delta }_{t-1}^{v}, & t > 1,{m}_{t-1}^{v}==0\\ 
        s_{t}-s_{t-1}, & t > 1,{m}_{t-1}^{v}==1\\ 
        \mathrm{0,} & t==1
    \end{cases}      
    \label{eqn:timelag}
\end{equation} \\

\noindent \textbf{Definition 4:} \textit{Longitudinal data imputation} \\
The goal of LDI is to address the problem of missing, in particular in the temporal data component. However, most methods aim to generate a complete time series $\hat{X}$. The imputation objective is then to estimate:
\[
\hat{x}_t^v = f(X, M, \theta),
\]
where $\theta$ is the set of model parameters and $f$ is the imputation function that seeks to minimise:
\[
\|\hat{x}_t^v - x_t^v\|^2, \quad \forall m_t^v = 0.
\]
where $\hat{x}_t^v \in \hat{X}$ and $m_t^v$ is defined in Eq. \ref{eqn:missmask}. \\

\noindent \textbf{Definition 5:} \textit{Longitudinal data classification} \\
For longitudinal data given in Definition 1, let $y_i\in\{1, \dots, K\}$ be the class label for $i$th instance. The classification objective is then to find:
\[
y=\arg\max_{y} P(y|X,\theta)
\]
where $\theta$ is the set of model parameters. In generative modelling, the objective is to maximise the likelihood of observing $P(X,y,\theta)$, thereby estimating a realistic posterior $P(y|X,\theta)$ to optimise classification performance. \\


\subsection{Generative Adversarial Networks} \label{sec:gans}
\begin{figure}[!ht]
    \centering
    \includegraphics[width=1.0\textwidth]{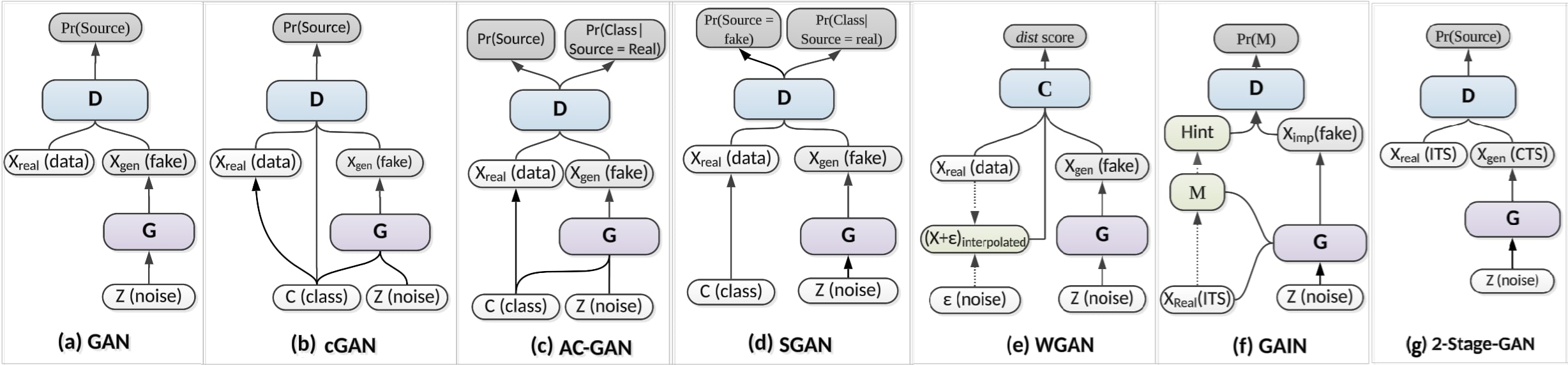}
    \caption{(a) Traditional GAN \cite{goodfellow2014gan}, (b) Conditional GAN (cGAN) \cite{mirza2014conditionalgans}, (c) Auxiliary Classifier GAN (AC-GAN) \cite{odena2017acgan}, (d) Semi-supervised GAN (SGAN) \cite{odena2016semigan}, (e) Wasserstein GAN (WGAN) \cite{arjovsky2017wgan}, (f) Generative Adversarial Imputation Network (GAIN) \cite{yoon2018gain}, and 2-Stage GAN (GRUI-GAN) \cite{luo2018gruigan}.}
    \label{fig:gans}
\end{figure}

This section introduces the traditional GAN and other GAN architectures often adapted to address missing value imputation in longitudinal data. GANs implicitly estimate the training data's distribution though sampling. While inherently unsupervised, GANs are trained with a supervised objective and can be used in semi-supervised and supervised learning tasks. 
Figure \ref{fig:gans} shows notable GAN architectures that have either inspired or served as a foundation for most GANs addressing longitudinal missing data. At its core, GAN typically comprises two deep-learning based models: a generator $G$ and a discriminator $D$ \cite{goodfellow2014gan}. The $G$ aims to learn the underlying data distribution in order to generate samples appearing authentic enough to be mistaken for real. The $D$ stifle $G$'s efforts by labelling all its generated samples as fake, its success is constantly fed back to $G$, which uses the information to improve its samples, until $D$ can no longer distinguish them from the training data set. The objective functions for each GAN's $G$ and $D$ are provided. 

The data $X$ in the following seminal GAN models can represent any type of data, e.g. images or text, and $y$ as data or embedding provided to condition the $G$. However, in terms of longitudinal data, $X$ here is taken mostly as the time series component $X^V_{1:T}$ defined under Section \ref{sec:definitions}, although it can mean $X_i$ to include the static component and/or class labels, which is what $y$ would represent for LDC tasks.   

\subsubsection{Traditional GAN \cite{goodfellow2014gan}} \label{sec:gan}
Shown in Figure \ref{fig:gans}(a), the original GAN model aims to balance both $D$ and $G$'s objectives by minimising the Jensen-Shannon Divergence (JSD) \cite{Tsai2005jsdistance} between the real data distribution $p_r$ from which the training data is observed, and $G$'s learned distribution $p_g$, from which fake samples are generated. $G$ utilises a prior noise distribution $p_z$, such as a Gaussian distribution, to learn a mapping from the latent space $Z$ to the data space $X$. The noise vectors $z$ add variability to $G$'s generated samples, otherwise $G$ would be merely replicating samples from $X$. $D$ trains as a classifier, to distinguish real samples from synthetic ones. The $G$ and $D$ objectives for the traditional GAN are expressed as:  
\begin{align}
   \mathbf{L}_{\text{G}}  &= \mathbb{E}_{z\sim p_z}\big[\log (1-D(G(z)))\big] \label{eq:gan_gloss} \\
   \mathbf{L}_{\text{D}}  &= \mathbb{E}_{x\sim  p_r}\big[\log D(x)\big]  + \mathbb{E}_{z\sim p_z}\big[\log (1-D(G(z)))\big] \label{eq:gan_dloss}
\end{align}
where $G(z)$ is the synthetic sample, and $D(G(z))$ and $D(x)$ represent $D$'s output probabilities, for whether a sample is fake or real. In essence, $D$ maximises the likelihood of correctly predicting the samples' source $Pr(Source)$, as shown in Figure \ref{fig:gans}, while $G$ minimises $\log (1-D(G(z)))$, the likelihood that $D$ is right about its samples' source. However, this approach can cause vanishing gradients early in training \cite{goodfellow2014gan}. Since $G$ relies on $D$'s error signals to improve, it can saturates quickly if $D$ becomes too proficient at distinguishing its poorly generated samples at initial stages of training \cite{arjovsky2017traingans}. To prevent this, $G$ often maximises $D(G(z))$ instead, which is the likelihood that $D$ misclassifies its generated samples. Both $G$ and $D$ continue to improve on their objectives until they reach Nash Equilibrium \cite{osborne1994nashequilibrium}, at which point the GAN model converges.

 In terms of longitudinal data, the $G$ and $D$ may be adjusted to take in both static and temporal data, as well as class labels, e.g. $D(X^Q,X^V,y)$. However, this would require careful integration of the data components to ensure the complex feature dependencies are captured.

\subsubsection{Conditional GAN (cGAN) \cite{mirza2014conditionalgans}} \label{sec:cgan}
Shown in Figure \ref{fig:gans}(b), cGAN was proposed to address mode collapse (see section \ref{sec:gan_challenges}) by enabling $G$ to learn a conditional distribution $P(X|y)$, where $y$ here can be class labels or any \enquote{extra} information provided \cite{mirza2014conditionalgans}. The cGAN objective function for $G$ and $D$ are expressed as:
\begin{align}
    \mathbf{L}_{\text{G}} & = \mathbb{E}_{z\sim p_z} \big [\log (1-D(G(z \mid \mathrm{y}))) \big] \label{eq:cgan_gloss} \\
    \mathbf{L}_{\text{D}} & = \mathbb{E}_{x\sim p_r} \big [\log D(x \mid \mathrm{y}) \big] \\
                          & + \mathbb{E}_{z\sim p_z} \big [\log (1-D(G(z \mid \mathrm{y}))) \big] \label{eq:cgan_dloss}
\end{align}
Here, class labels $\mathrm{y}$ fed to $G$ are synthetic, as $G$ does not access real data. The \enquote{conditioning} can also come from other parts of the real data \cite{Bernardini2023ccgan}, other modalities \cite{hyland2018cgants}, outputs from other models like autoencoders \cite{zhang2021e2ganrf}, from static features in longitudinal data, or even from $G$ itself in recurrence-based GANs for imputation, where the outputs are fed back into the model \cite{ma2020ajrnn}. 

\subsubsection{Auxiliary Classifier GAN (AC-GAN) \cite{odena2017acgan}} 
Illustrated in Figure \ref{fig:gans}(c), AC-GAN improves cGAN \cite{mirza2014conditionalgans} by adding a classification objective to $D$. Unlike cGAN, AC-GAN does not feed class labels to $D$ as input; instead, $D$ must learn this \enquote{side} information from $G$ to help generate class-conditional samples. AC-GAN uses fully labelled data, with $D$ also outputting class predictions on the real data. 

$D$ has two output arms: its auxiliary classifier arm $\text{D}_C$ decodes class labels from the generated data and outputs class probabilities ($P(class)$); and its source discriminator arm $\text{D}_S$ outputs source probabilities ($P(Source)$). These objectives are optimised through $D$'s class discrimination loss $\mathbf{L}_{\text{D}_C}$ and source discrimination loss $\mathbf{L}_{\text{D}_S}$, respectively. AC-GAN's objective functions are defined as:
\begin{align}
    \mathbf{L}_{\text{G}} & = -\mathbb{E}_{z\sim p_z}\big[\log \text{D}_S(G(z \mid \mathrm{y}))\big] + \log \text{D}_C(y \mid G(z \mid \mathrm{y}))\big] \\
    \mathbf{L}_{\text{D}_{S}} & = \mathbb{E}_{x\sim p_r}\big[\log \text{D}_S(x)\big]
                            + \mathbb{E}_{z\sim p_z}\big[\log (1-\text{D}_S(G(z \mid \mathrm{y}))\big]\notag\\ 
    \mathbf{L}_{\text{D}_C} & = \mathbb{E}_{x\sim p_r}\big[\log \text{D}_C(\mathrm{y} \mid x))\big]
                            + \mathbb{E}_{z\sim p_z}\big[\log (1 - \text{D}_C(\mathrm{y} \mid G(z \mid \mathrm{y}))\big]\notag\\ 
    \mathbf{L}_\text{D} & = \mathbf{L}_{\text{D}_S} + \mathbf{L}_{\text{D}_C}
\end{align}
where class labels $y$ taken in by $G$ are also fake. GAN-based models with an added classification objective to the main LDI objective may implement $\text{D}_S$ and $\text{D}_C$ as a multi-task learning model \cite{caruana1997multitasklearning,hwang2018imputation}, as separate models  \cite{miao2021ssgan} or with classification model $\text{D}_C$ only \cite{huang2021itgan}.
 
\subsubsection{Semi-supervised GAN (SGAN) \cite{odena2016semigan}} 
Referred to sometimes as SSL-GAN, the SGAN model in Figure \ref{fig:gans}(d), enhances the quality of generated samples $G(z)$ by utilising partially labelled data. Here, $D$ is trained to classify samples into one of $(K+1)$ conditional distributions: $\{P(y=1|X), ... P(y=K|X)\}$ for $K$ real classes, and $P(y=K+1|G(z))$, where fake samples are assigned to the $(K+1)$th class. 

Based on the observed samples, $D$ outputs probabilities for each class-based distribution. The unlabelled portion of the data $\mathcal{U}$ is used to train $D$ in an unsupervised manner, distinguishing between real samples $P(y \leq K | X)$ and fake samples $P(y=K+1|(G(z))$, based on an \textit{unsupervised} loss $\mathbf{L}_{\text{unsup}}$. Meanwhile, the labelled data $\mathcal{L}$ allows $D$ to minimise a \textit{supervised} loss $\mathbf{L}_{\text{sup}}$, helping $G$ produce realistic class-based samples. The overall SGAN objective is defined as:
\begin{align}
     \mathbf{L}_{\text{G}} & = -\mathbb{E}_{z\sim p_z}\big[\log D(y\leq K|G(z))\big] \\
     \mathbf{L}_{\text{unsup}} & = -\mathbb{E}_{x\sim \mathcal{U}}\big[\log(1-D(y = K+1|x))\big]  + \mathbb{E}_{z\sim p_z}\big[\log D(y=K+1|G(z))\big]  \notag\\ 
     \mathbf{L}_{\text{sup}} & = \mathbb{E}_{x\sim \mathcal{L}}\big[\log D(y|x,y\leq K)\big] \notag\\     
     \mathbf{L}_{\text{D}}  & = \mathbf{L}_{\text{unsup}} + \mathbf{L}_{\text{sup}}      
     \label{eqn:ssganloss}
\end{align}
A concurrent work \cite{salimans2016semisupervisedgans} proposes the use of feature-matching loss $\mathbf{L}_{\text{FM}}$ in SGANs to replace the $G$ loss:
   $\mathbf{L}_{\text{FM}} = \mathbf{L}_{\text{G}} = \left\| \mathbb{E}_{x \sim \mathcal{U}} \mathbf{f}(x) - \mathbb{E}_{z \sim p_z} \mathbf{f}(G(z)) \right\|_2^2$
where $f(^*)$ are outputs of intermediate layers of $D$ that allow $G$ to learn most discriminative features of the real data only, thus stabilising training. Also, class-specific longitudinal sequences can be generated give class labels.

\subsubsection{Wasserstein GAN (WGAN) \cite{arjovsky2017wgan}}
Shown in Figure \ref{fig:gans}(e), WGAN is aimed at stabilising training in GANs by mitigating mode collapse and exploding/vanishing gradients. WGAN optimises the Wasserstein loss (W-loss) based on the Earth Mover's distance \cite{frogner2015wloss}. Here, $D$ is replaced by a Critic $\mathbb{C}$ that outputs a real-valued distance score indicating how far apart $p_r$ and $p_g$ samples are. This is the adversarial loss, denoted here as $\mathbf{L}_{\text{critic}}$ that $\mathbb{C}$ aims to maximise and $G$ tries to minimise.
 
Unlike GAN, WGAN's outputs are not passed through a $log$ function, making $\mathbb{C}$ unbounded and 
preventing vanishing gradients. To ensure valid W-loss, WGAN must satisfy 1-Lipschitz (1L) continuity,
which is achieved through methods like weight-clipping \cite{arjovsky2017wgan}, spectral normalisation \cite{miyato2018specnorm} and gradient penalty (GP) \cite{gulrajani2017wgangp}. With a GP term $\mathbf{L}_\text{penalty}$ that penalises $\mathbb{C}$ if the gradient norm deviates from 1.0, the WGAN's objectives with GP are given below:
\begin{equation}
    \begin{split}
        \mathbf{L}_\text{G} & = - \mathbb{E}_{z\sim p_z}\big[\mathbb{C}(G(z))\big]\\
        \mathbf{L}_\text{critic} & = - \mathbb{E}_{x\sim p_r}\big[\mathbb{C}(x)\big] + \mathbb{E}_{z\sim p_z}\big[\mathbb{C}(G(z))\big]\\
        \mathbf{L}_\text{penalty} & = \mathbb{E}_{\hat{x}\sim p_{\hat{x}}}\big[(||\nabla \mathbb{C} (\hat{x})||_{2}-1)^2\big] \text{, where } \hat{x}  = \epsilon \cdot x - (1-\epsilon) \cdot G(z)\\
        \mathbf{L}_\mathbb{C} & = \mathbf{L}_\text{critic} + \lambda \mathbf{L}_\text{penalty}
        \label{eqn:wgan}
    \end{split}
\end{equation}
where $\lambda$ is a regularisation parameter, 
$\hat{x}$ is an interpolation of real $x$ and fake $G(z)$ samples, helping $\mathbb{C}$ learn between real and generated data, and $\nabla$ is $\mathbb{C}$'s gradients with respect to $\hat{x}$. 

\subsubsection{Generative Adversarial Imputation Network (GAIN) \cite{yoon2018gain}}
Shown in Figure \ref{fig:gans}(f), GAIN is a popular GAN for imputation of missing tabular data. The $G$ takes three inputs: real data $X$ with missing values $\tilde{X}$, the missing value mask $M$ in Eqn. (\ref{eqn:missmask}), and noise $Z$ for seed estimates of missing values. $G$'s output $\Bar{X}$ and the final imputed data $\hat{X}$ are given as:
\begin{align}
    \Bar{X} & = G(\tilde{X},M,(1-M)\odot Z) \label{subeqn:xgen} \\
    \hat{X} & = M \odot \tilde{X} + (1-M)\odot \Bar{X} \label{eqn:x_hat} 
\end{align}
The imputed data matrix $\hat{X}$ includes $G$'s generated estimates for missing values.  
$D$ evaluates each element of $\hat{X}$, determining whether it is real or imputed (fake) by reproducing the missingness mask $M$.

To guide $D$, GAIN introduces \enquote{Hints} $H$, defined as:
\begin{equation}
    H = \beta \odot M + 0.5(1-\beta)
    \label{eqn:hint}
\end{equation}
where $\beta$ is a random binary variable controlling which elements of the mask $M$ are revealed to $D$. $H$ ensures $G$ learns the true data distribution, given the missing data distribution \cite{yoon2018gain}. The $G$ objective of GAIN is composed of its adversarial objective function and a reconstruction loss $\mathbf{L}_{\text{rec}}$. GAIN's losses are given as below: 
\begin{align}
    \mathbf{L}_\text{rec} & = \mathbb{E}_{\bar{X},\tilde{X},M} \big[ M^T (\tilde{X} - \Bar{X})^2 \big] \\
    \mathbf{L}_\text{G} & = -\mathbb{E}_{\hat{X},M,H} \big[ {(1-M)^T}\log{D({\hat{X} },H)}\big] + \lambda \mathbf{L}_\text{rec} \\
    \mathbf{L}_{\text{D}} & = \mathbb{E}_{\hat{X},M,H} \big[ M^T\log{D({ \hat{X} },H)} +{(1-M)^T}\log{(1-D({ \hat{X} },H))} \big] \\
        \label{eqn:gain-loss}
\end{align}
where the $G$ is improved with a weighted reconstruction loss.

\subsubsection{Gated Recurrent Unit with Imputation in GANs (GRUI-GAN) \cite{luo2018gruigan}} \label{sec:gruigan}
Also known as the two-staged GAN (Figure \ref{fig:gans}(g)), GRUI-GAN was one of the first GAN models for time series imputation. It proposed a modified GRU cell for imputation (GRUI). GRUI-based GANs leverage temporal decay rates (Eq. \ref{eqn:timelag}) to control the influence of past observations on imputation estimates, where longer intervals have higher decay rates, reducing their impact.

In GRUI-GAN, designed specifically for time series imputation is performed after GAN training, referred to here as \enquote{post-training} imputation. The imputed sample $\hat{X}$ is gained similarly as in Eq. \ref{eqn:x_hat}.
GRUI-GAN's $G$ is optimised using a combination of WGAN loss and a mean squared error (MSE)-based reconstruction loss $\mathbf{L}_{\text{rec}}$, which ensures consistency between the observed and reconstructed time series. The discriminative loss from $D$ measures the quality of the generated samples. The GRUI-GAN losses are defined below:
\begin{align} 
    & \mathbf{L}_\text{rec} = \parallel X \odot M-G(Z)\odot M \parallel_{2}
    \label{eqn:imploss_gruigan}\\
    & \mathbf{L}_\text{G} = - \mathbb{E}_{z \sim p_z} [D(G(Z))] + \lambda \mathbf{L}_\text{rec}  
    \label{eqn:gloss_gruigan}\\
    & \mathbf{L}_\text{D} = - \mathbb{E}_{x \sim p_r} [D(X))] + \mathbb{E}_{z \sim p_z} [D(G(Z))]  
    \label{eqn:dloss_gruigan}
\end{align}

While GRUI-GAN has gained popularity, it faces challenges like error accumulation over time due to its dependence on GRUIs, especially as missing values increase \cite{bengio2015biasexploding}. Methods like BRITS's next-step prediction \cite{cao2018brits} help to mitigate this 
problem, as discussed in Section \ref{sec:gan_challenges}.

\subsection{The GAN-based Approach for LDC with Missing Data} 
\label{sec:ldc_gans}
\begin{figure}
    \centering
    \includegraphics[width=0.75\linewidth]{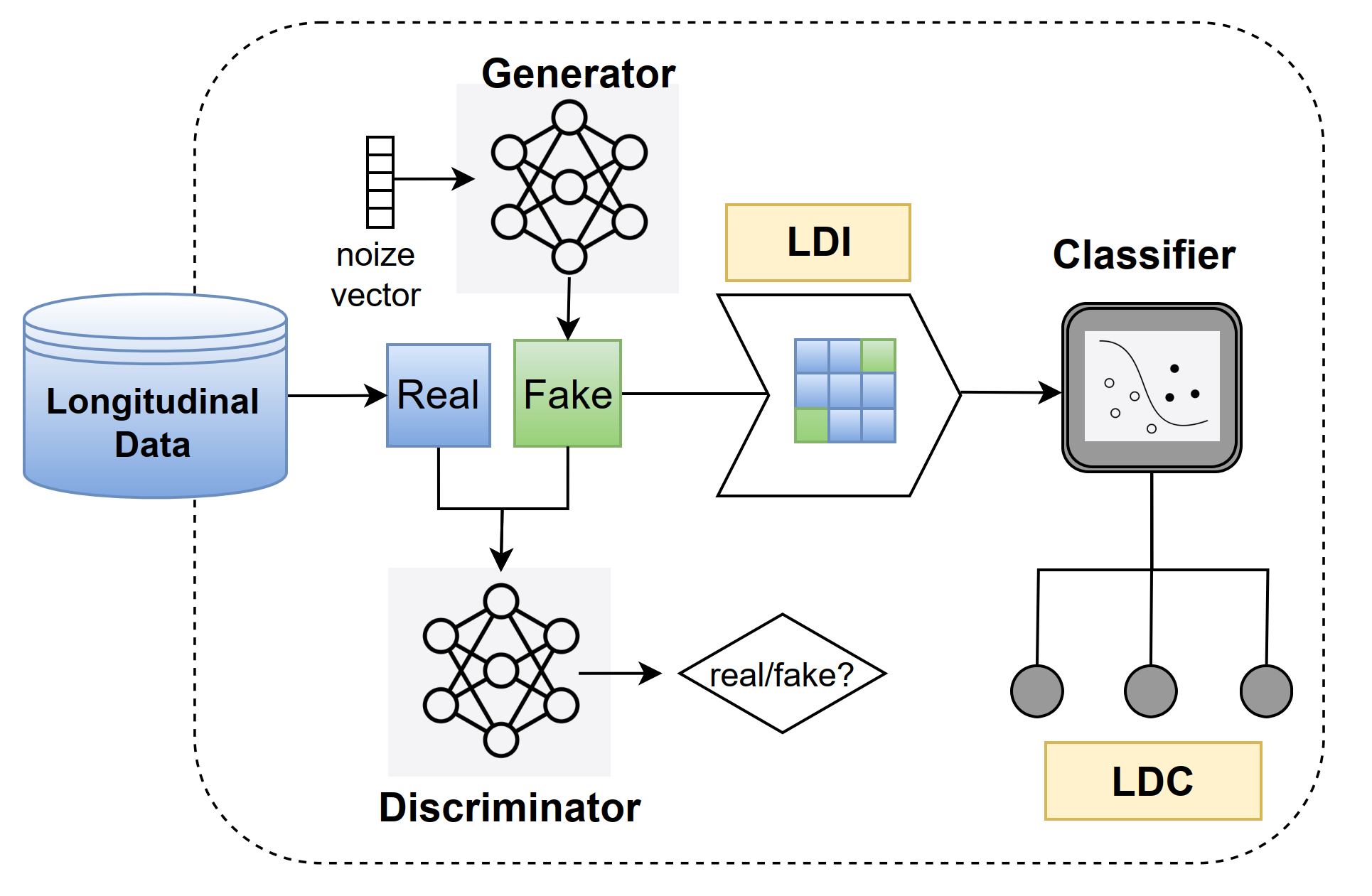}
    \caption{A two-phase GAN framework for LDC showing the input into the traditional GAN (with Generator and Discriminator networks) for feature learning and sample generation, imputation of missing values (LDI), and subsequent classification (LDC).}
    \label{fig:framework}
\end{figure}

There are two primary approaches for using GANs in LDC with missing data, the more common two-phased model approach (see simplistic outline in Figure \ref{fig:framework}, and the end-to-end joint models for imputation and classification. Research has focused primarily on the temporal component of longitudinal datasets, neglecting multi-dimensional nature of the data and its inherent heterogeneity. 

\subsubsection{Two-phase approaches} \label{sec:two-stage}
This approach encompasses the majority of research on GAN-based LDI, where imputation is performed as a preprocessing step and LDC is subsequently applied to evaluate the quality of the imputed samples. Initially, models in this category first aimed to learn the underlying data distribution to generate realistic samples, and then optimise for imputation quality \cite{luo2018gruigan}. However, optimising the noise vector and then optimsing for imputation is less effective \cite{luo2019e2gan} than directly generating samples that closely resemble the original time series for imputation. Importantly, as missing values may not respect the joint distribution of features in the data \cite{yoon2018gain}, require additional modeling assumptions \cite{little2019missing}, and their patterns can be informative for analysis tasks \cite{diggle1994informative}, jointly optimizing LDI and LDC objectives is likely to yield superior results.

\subsubsection{Joint end-to-end approaches} \label{sec:joint-approach}
This approach involves GAN-based LDI and LDC, where a classification objective is integrated into a GAN for LDI, either as an auxiliary task or as a sub-task within the GAN model. This approach has been shown to improve the primary task such as imputation \cite{kang2021rebootingacgan}; leading to mutual performance improvement for both tasks \cite{odena2017acgan}. Such GAN settings may have a separate classifier to the GAN \cite{deng2021ibgan,ma2020ajrnn}, or the discriminator can have a multi-task objective - adversarial and classification \cite{cui2020conan,hwang2018imputation}. 
 
\section{Methodology of survey}  
\label{sec:survey_methodology}
We follow the systematic approach of the PRISMA framework \cite{page2021prisma} to identify relevant literature. We crafted a search strategy based on our research questions to identify relevant papers across five digital databases. 

\subsection{Research questions}
Our primary focus is on methods and approaches that address the missing data challenge in longitudinal data to improve LDC. Although not all GAN-based LDI works aim to improve classification, most seek to improve the dataset quality, indirectly supporting LDC. We also focus on how these methods handle the data-level challenges presented by longitudinal data, especially in their treatment of static features, and the learning of temporal data given its sparsity. Our research questions are:
\begin{enumerate}
    \item \textit{What data-level challenges affect LDI for better classification outcomes?} \\
    If neglected, data-level challenges can lead to biased outcomes, misleading conclusions, loss of contextual information, and difficulty in learning patterns. This question helps us categorise the challenges and how they are addressed in GANs-based LDI methods.
    \item \textit{What LDI approaches have been developed utilising GANs?} \\
    GANs for LDI must first generate realistic context-aware longitudinal samples for imputation. We identify efficient GAN-based methods that support LDI and LDC. 
    \item \textit{How data-level challenges like instance heterogeneity, class imbalance and mixed data types have been addressed in GAN-based LDI methods?} \\
    Although there has been significant work on GANs for LDI, there is a need for theoretical and experimental evaluations of how broader data-level challenges, like the data's multi-dimensionality, are incorporated into the imputation process. 
\end{enumerate}

\subsection{Literature selection \& search Strategy}
We searched Scorpus, IEEExplore, Web of Science, ACM Digital Library and PubMed for works on GANs applied to LDI and LDC. Our synthesis covers papers from the last five years (2019-2023), including the current year's works. We also review foundational GAN papers that shaped the field, particularly in time series and longitudinal data imputation. Initial search for the keyword \enquote{Generative Adversarial Network} yielded about 130K articles, while the keywords [\enquote{missing values} OR imputation] returned over 56K, and [longitudinal OR \enquote{time series}] returned about 11.5M records. 

To refine the search, we used the following string ((\enquote{Generative Adversarial Net*} OR \enquote{Generative Adversarial Imputation} OR GAN) AND (imputation OR \enquote{missing data} OR incomplete) AND (longitudinal OR \enquote{panel data} OR \enquote{time series} OR multivariate OR health OR medical OR EHR) AND NOT (images OR scans OR MRI OR spatio)). On June 7, 2024, this search returned 223, 219, 62, 27 and 8 documents from Web of Science, Scopus, PubMed, IEEExplore and ACM Digital Library, respectively. The focus on time series data is due to the frequent use of missing data imputation in the time series component of longitudinal datasets, while terms from the medical domain were included because of the longitudinal nature of health-related datasets.

\subsection{Inclusion \& exclusion criteria}
To narrow the large number of results, we used the following criteria to select relevant studies:
\begin{enumerate}
    \item peer-reviewed
    \item in the English language
    \item available online and in full-text
    \item published in the last five years (2019 - 2024)    
    \item multivariate data, excluding univariate time series data.
    \item focused on GAN-based LDI, including hybrid models like GANs with autoencoders. 
    \item LDI related only, 
    excluding pure time series studies 
    \item focused on structured or tabular data, excluding longitudinal images, spatio-temporal data or token sequences like medical codes. 
\end{enumerate}

Foundational work on GANs for time series and longitudinal data synthesis began in 2018, particularly within the medical field. While we draw on these works to analyse trends, we primarily focus on post-2018 papers where GAN-based LDI gained traction.

\subsection{Item extraction \& synthesis}
After removing duplicates and applying the inclusion and exclusion criteria, there remained 56 articles for review. We synthesise the papers based on:
\begin{enumerate}
    \item  Approaches used in GANs for LDI to address data-level challenges in LDC (see Figure \ref{fig:topolgy_challenges}), and how methods have evolved over the last 5 years.
    \item Gaps and challenges that the GAN-based LDI methods addressed 
    \item Common techniques and innovative solutions developed to improve LDI with GANs. 
\end{enumerate}
In accordance with our research questions, we first present the data-level challenges in LDC, the assumptions around longitudinal data, and their implications for LDC outcomes. 

\section{Challenges in LDI \& LDC with GANs} 
\label{sec:LDC_GAN_challenges}
\begin{figure}[!ht]
    \centering
    \includegraphics[width=1.0 \textwidth]{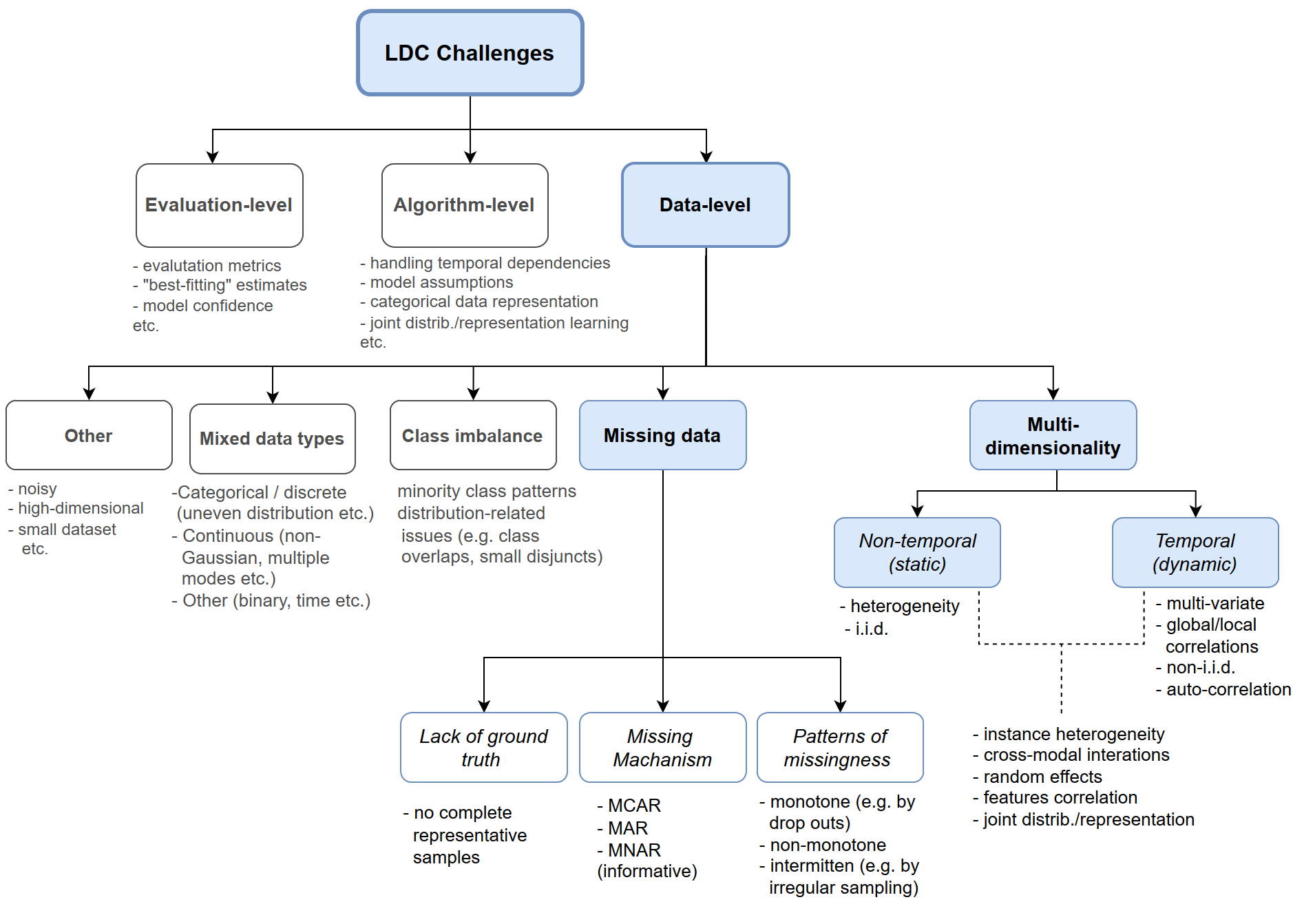} 
    \caption{A topology of LDC Challenges faced by GANs when modelling longitudinal data, for improving tasks like LDC. The highlighted boxes represent the most critical challenges for longitudinal data imputation discussed in-depth in this survey.}
    \label{fig:topolgy_challenges}
\end{figure}

The multi-dimensional nature of longitudinal data introduces several layers of complexity to classification tasks, particularly in the joint modelling of static and temporal components. While this paper focuses on using GANs for imputing missing data for improved LDC, it is important to acknowledge the broader challenges posed by combining the static and temporal data elements for LDC. 
These data-level challenges, shown in Figure \ref{fig:topolgy_challenges}, affect any method aimed at improving the quality and utility of longitudinal data, whether for LDC or other analysis tasks.

We discuss here data-level and GAN-based challenges to longitudinal data imputation and classification. In particular, we focus on Additionally, we discuss other inherent challenges, such as instance heterogeneity from static components and both local and global temporal correlations in the temporal data. 

\subsection{Missing data challenge}
\label{sec:challengs_mvs}
The complexity of dealing with missing longitudinal data has to do with inter-related aspects like the underlying mechanism causing the missingness, the resulting patterns of missing data, and the absence of ground truth for validation. The missingness mechanism or assumption, whether it is Missing At Random (MAR), Missing Completely At Random (MCAR), or Missing Not At Random (MNAR) \cite{rubin1976missingdata} influences observed patterns of missingness in the data and the ability to validate LDI methods. For instance, monotone missingness patterns from study drop outs may indicate MNAR \cite{enders2011missing}, where the missing data itself is informative to the analysis task \cite{diggle1994informative}. This relationship between mechanism and pattern directly impacts the availability of complete records for validation, as certain mechanisms may systematically prevent the collection of \enquote{ground truth} data. Studies \cite{sun2020irregular} have shown missing rates to exceed $80\%$ in popular longitudinal datasets like PhysioNet, where irregular sampling contributes up to $60\%$ of missing data; of course, rates $>15\%$ can significantly impact classifier performance \cite{acuna2004missing}. A detailed understanding of these challenges is essential for developing effective imputation strategies.

\subsubsection{Missingness mechanisms.}
Missing values in longitudinal data are primarily caused by irregular sampling and drop out, although measurement errors can contribute. Irregular sampling affects temporal alignment, where time series observations across and within instances do not share a common timeline with matching time points of observations \cite{Pei2021generatets}. Drop outs are particularly common in human studies, where participants discontinue partway, resulting in partial time series data. The underlying mechanism to the missing data determines the effectiveness of imputation methods \cite{little2013missing}, including those GAN-based. 

Typically, a MAR mechanism is assumed for systematic missingness like irregular sampling rates \cite{wood2019comparing}, where observed data can explain missing values. That is, where the probability of missingness $P(M|X)=P(M|X_{\text{obs}})$ depends on the observed data $X_{\text{obs}}$. $X$ represents the complete data (observed and missing) and $M$ is the missingness indicator. If the probability of missingness is independent of the data $X$, as in the case of random missing values, then data is MCAR, and $P(M|X) = P(M)$. In contrast, if the probability of missingness depends on the missing data itself, it is considered MNAR \cite{frees2004longitudinal} and given as $P(M|X) = P(M|X_{\text{obs}}, X_{\text{mis}})$, where $ X_{\text{mis}}$ is the missing data. 

While statistical tests like Little’s MCAR test \cite{wang2023score,little1988test}, and the multiple imputation-based sensitivity test \cite{beesley2021multiple} exist to detect mechanisms present, often, the difference between MAR and MNAR require domain knowledge and insight into the dataset. E.g. missing end-point data resulting from study drop outs where mortality is tracked among patients, would imply MNAR, with missingness patterns informative to the classification task \cite{zhang2020informdropout}, and relevant to observations taken over time. Auxiliary data, such as static features, can provide some indication of the missing data mechanism. For example, if it is deducted that female participants are more likely to have missing salary information, then it MAR would result. 

MAR is the most common assumption in practice, and estimates for missing values benefit from multiple imputation, as used in GANs  \cite{awan2021imputationcgain}. As most longitudinal data, particularly EHRs are considered MAR \cite{little2019missing}, experts \cite{demirtas2018flexibleimputation} suggest beginning with the MAR assumption, using multiple imputation techniques and testing for robustness under different mechanisms. MNAR is non-ignorable and the most difficult to address, it is commonly assumed for irregularly sampled data with drop outs. Imputation methods for MNAR require explicitly modelling the missing data patterns and often require external validation \cite{che2017grud,demirtas2018flexibleimputation}. 

\subsubsection{Lack of ground truth}
A significant challenge in handling missing longitudinal data is the absence of ground truth. This lack of validation data arises from inherent missing data mechanisms and resulting missing data patterns. In datasets with high levels of data missingness, such as PhysioNet \cite{sun2020irregular}, complete records for validation are often unavailable, pose significant challenges for both imputation and classification \cite{che2017grud}. For instance, study drop outs cause the loss of end-point data, which prevent accurate recovery of temporal dependencies. 
Explicitly incorporating missingness information can help mitigate this issue by modeling the informativeness of missing data patterns \cite{Lipton2016missingrnns}.

The absence of ground truth emphasises the importance of robust imputation strategies. GAN-based methods have shown promise by leveraging adversarial training to approximate realistic imputations. Integrating data missingness patterns and mechanisms into the imputation and modelling process are essential to ensure accurate validation and classification performance.

\subsubsection{Missing data patterns}
Often, meaningful patterns exist in missing data \cite{qian2024howdeep}. For example, medical data is often irregularly sampled and temporally unaligned, while education data may be regularly sampled and temporally aligned \cite{shukla2020surveyirregts}. Drop outs display mono-tone patterns where data is systematically missing beyond a certain point in time \cite{wood2019comparing}. This type of missingness causes temporal patterns beyond the drop out to be lost, and can impact imputation. If the drop out is assumed to be MNAR, then this dependency must be explicitly modelled \cite{che2017grud}. Irregular sampling, under a discretised representation \cite{shukla2020surveyirregts}, displays intermitted patterns of missingness along evenly spaced time points. Although this representation enforces temporal alignment, it often results in highly sparse datasets with a lack ground truth, where complete representative samples are unavailable for model validation purposes. Patterns of missingness can be particularly useful for imputation under such circumstances \cite{yoon2018gain}. 

Works circumventing the missing data problem by omitting incomplete data cause classifiers to wrongly assume that the remaining complete data is representative of the entire dataset \cite{rubin1976missingdata}. This and other unrealistic assumptions regarding data missingness can lead to biased parameter estimation and reduce model accuracy \cite{dong2013missing}. Techniques such as interpolation have often been used to restore temporal alignment and capture intra-instance correlations, particularly when analysing the progression of an outcome of interest \cite{marti2020lng}. However, studies that do not regard irregularly sampled data missing values often represent the MTS as time-value pairs or a collection of univariate time series (see Definition 2 in Section \ref{sec:bg_ldc_defns}) \cite{guo2019mtsgan}. Although this view avoids the introduction of missing data, it violates the non-i.i.d. assumption between temporal features coming from the same instance. While we discuss irregularly sampled longitudinal data under the missing values problem, we acknowledge methods that address sampling irregularity outside of the missing data problem space.

\subsubsection{Handling missing data in GANs.}
While discriminative methods struggle to handle missing data directly, GANs can marginalise out missing data by factoring the joint distribution below:
\begin{align*}
    P(X_{\text{obs}}, M) & = \int P(X_{\text{obs}}, X_{\text{mis}}, M) \, dX_{\text{mis}} \\
    & = P_{\theta}(X_{\text{obs}})P_{\phi}(M|X_{\text{obs}})
\end{align*}
where the correct estimation of $\theta$ renders the missing data mechanism inconsequential in non-temporal datasets when generating $P_{\theta}(X_{obs})$ \cite{li2019misgan}. However, this approach becomes problematic in longitudinal data, given instance heterogeneity and temporal correlations, where marginalising out missingness from irregular sampling can distort missing value estimations. Incorporating the missingness pattern information assists in factoring the informativeness of missingness distribution \cite{diggle1994informative}. Another situation of lacking ground truth, is when end-point data (latter time steps) are missing, and only prefix (earlier time step observations) available for analysis, like early LDC.

\subsection{Multi-dimensionality challenge}
Imputation methods that assume a joint distribution over the data, as in longitudinal data, may face challenges due to complex dependencies and temporal correlations that can lead to biased or distorted imputation estimates. As shown in Figure \ref{fig:topolgy_challenges}, both static and temporal components present their unique challenges. Static features capture baseline characteristics of the population and account for some random effects often considered in mixed models  \cite{ngufor2019longitudinalmixedeffects}. 
These features cause {instance heterogeneity}, variations within the instances' features over time that can influence predictions \cite{fitzmaurice2012long,verbeke2010long}. Ignoring the correlations between static and temporal features may lead to models that overfit 
\cite{sheetal2023longdatabestpractice}. 
The time series component contains local (intra-series or within-instance) and global (inter-series or between instance) correlations that must be considered for accurate modeling \cite{wang2023local}. Auto-correlation or serial correlation within an instance \cite{zhou2018long} measures the influence of previous observations on the current one, but irregular time gaps can disrupt these patterns, requiring thoughtful imputation. 


\subsection{Other challenges} Following are some other challenges.

\textbf{Class imbalance.} When certain classes have significantly less instances than others, class overlap and small disjuncts are introduced, which weaken the classifier's discriminative performance \cite{fernandez2018imbalance}. GANs are often used to address this \enquote{small data size problem} by generating synthetic minority class data \cite{pan2024ganimbal}. Assessing the diversity and novelty of these synthetic samples is necessary. In terms of LDI or LDC, deep learning models, including GANs, often assume balanced class distributions \cite{johnson2019imbalance}. However, real-world datasets frequently exhibit imbalances, particularly in critical domains like health. When minority classes are underrepresented, GANs, which perform best with diverse, large datasets, may overlook them entirely \cite{krawczyk2016imbalance}. Mode collapse and convergence issues arise, the effects of which on LDI are discussed in Section \ref{sec:gan_train_challenges}. GANs have been adapted for imbalanced learning in supervised tasks \cite{deng2021ibgan,ma2020maskehr}. Oversampling and semi-supervised learning approaches have been used for detecting rare classes \cite{sampath2020surveygan}, e.g. rare disease identification in healthcare \cite{cui2020conan,yu2019rarediseasessgan}. These approaches generate new data points through non-linear interpolation and utilise labelled and unlabelled data to improve imbalanced classification \cite{sampath2020surveygan,iwana2021augmentfortsclass}.

\textbf{Small dataset.} Many real-world longitudinal datasets are small, leading to potential overfitting in GAN-based models \cite{caiafa2021smalldataset}. 
Small datasets can cause $D$ to overfit to the training set  \cite{karras2020smalldatagan}. This leads training to diverge \cite{arjovsky2017traingans} and the generated samples to resemble the real data closely but lack diversity. Conversely, under-fitting produces diverse but less authentic samples \cite{adlam2019overunderfitgan}. While extensive research exists on data synthesis for longitudinal data, much of it relies on large-scale datasets \cite{choi2017medgan,weldon2021ganehr}. The standard GAN-based solution to over-fitting in image and other data types is data augmentation \cite{tran2020dataaugment}, though research on augmenting time series data for improve missing data imputation is still emerging. Particularly as imputation quality suffer because of the lack of ground truth available. 

\textbf{Mixed data types.}
Longitudinal data often include mixed types, such as numerical (discrete and continuous), categorical data and other types like dates. 
Given that deep learning-based methods like GANs can only take in continuous input, proper encoding methods are essential to preserve meaning and relationships between different data types. In terms of LDI, GANs would need to simultaneously impute continuous distributions and discrete categories of data. Given the challenge that different data types would require different imputation techniques and loss function \cite{khan2022mixedimputgan}, most research are focused on individual types, for instance for categorical data only \cite{yang2019categoricalgain}. 
Simple techniques like one-hot encoding for categorical data can lead to information loss and fail to capture complex interactions. GANs have addressed this by using feature embedding \cite{yoon2019tsgan}, conditioning the learning space with categorical information like class labels \cite{mirza2014conditionalgans} and the use of multiple encoders \cite{li2023ehrgenmixeddata}. 

\textbf{High dimensionality.} 
Some longitudinal datasets, such as patient genomics and finance data, are high dimensional, with more features than observations. This creates challenges in LDC due to the \enquote{curse of dimensionality}, increased computational complexity, and difficulties in selecting relevant features, particularly for contextual imputation, and introduce noise. GANs have shown promise in tackling these issues by effectively modelling and synthesising high-dimensional data, capturing complex patterns across numerous features \cite{lee2023ganshighdim}.


\subsection{GAN-based challenges} \label{sec:gan_challenges}
Given the complexity of GANs and training dynamics, they face multiple challenges regardless of the data type or specific task. These challenges stem from the model's basic assumptions regarding implicit representation learning and its training process.

\subsubsection{Joint distribution learning challenges}
The main objective of a GAN is to learn the underlying data distribution to enable data generation or sampling. This involves mapping a latent space, $p_z$, to a data space, $p_g$, that approximates the real distribution $p_r$. For this, GAN assumes that both the $G$ and $D$ have sufficient capacity in terms of design, complexity, and parameters to model the data generation function. Additionally, successful training requires large and diverse training datasets to capture $p_r$ \cite{karras2020smalldatagan}. Moreover, correctly specified learning objectives are essential to ensure convergence of $p_g$ to $p_r$ at Nash Equilibrium.

Some major challenges arise: first, the multi-dimensional nature of longitudinal data requires effective joint modeling of static and temporal variables, how they are distributed together, focusing on their relationships and dependencies. Second, the view of the longitudinal components as modalities \cite{Zhang2024pregating} require effective integration into a shared feature space for discriminative feature extraction for improved LDI and LDC. Finally, the sparsity of temporal data complicates the extraction of temporal relationships, particularly when missing data undermines temporal correlations.  

In practice, beyond the multi-dimensional challenges, these datasets are often small, noisy, irregularly sampled, imbalanced, contain missing values, and lack overall quality. Also, $G$ and $D$ may not always be able to avoid under or overfitting the data; hence, aligning model capacity with the task complexity is crucial \cite{Yang2022dyndiscgan}. 

\subsubsection{GAN training challenges} \label{sec:gan_train_challenges}
GANs are notoriously known for severe instability issues, which significantly impacts both the training process and the quality of the generated samples \cite{wiatrak2019ganstabilising}, in particular for tasks such as imputation. These are discussed particularly in the context time series generation and imputation, under which most research for LDI is concentrated. 

\textbf{Mode collapse} 
is a \enquote{lack of diversity} problem. In the general case, $G$, failing to capture the diversity of the underlying distribution, repeatedly produces a narrow variety of outputs. This causes the $D$ to become stuck in a local minima. For time series or longitudinal data generation and imputation, the $G$ may produce certain repetitive or homogeneous samples, which maybe relevant to specific population subset or time periods only. Missing data corrupts temporal dependencies, which would decrease reliability of generated estimates, particularly only dominant feature correlations are captured and vital but rare temporal patterns are missed \cite{tang2020lgnet}. Imbalanced classification tasks may fail to generalise well, given that minority classes or \enquote{modes} are ignored. Methods proposed to address these challenges include 1) class-conditional GANs \cite{zheng2016mccnn} where $G$ is forced to generate samples from all classes, 2) alternative loss functions like W-loss \cite{he2023transgan} and least-squared loss \cite{mao2017lsgan}, 3) Bayesian GANs \cite{Saatci2017bayesiangan}, 4) $G$ ensembles for each mode \cite{zhang2018stackelbergganmed}, and 5) others, such as including a memory module to learn global correlations in the data \cite{tang2020lgnet}. 

\textbf{Vanishing gradients}
occur when $D$ significantly outperforms $G$ early in the training. This causes insufficient feedback for $G$' improvement. For LDI or time series imputation, longer sequence lengths as well as temporal dependencies impact on gradient flow at back propagation. This collapse in training can degrade imputation estimates for missing data and subsequently degrade classifier performance. Solutions within GAN-based LDI literature include 1) W-GANs \cite{Festag2023rcgan}, where $G$ continues to receive meaningful feedback from the critic $\mathbb{C}$, which improves linearly throughout training, 2) a modified minimax loss function \cite{creswell2018ganoverview}, where 
changing $G$'s objective to $-log(D(G(z)))$ alleviates the vanishing gradient problem, 3) employing leakyReLu, which avoids small gradients 
typical of Sigmoid, TanH and ReLU. 
to be back propagated to $G$, 4) moment matching methods \cite{dziugaite2015momentmatchinggan} with loss functions like Maximum Mean Discrepancy (MMD) \cite{gretton2008mmd,Thanaraj2020gasf}, and 5) architectural redesigns like skip connection \cite{song2023dagan}. 

\textbf{Failure to converge}
occurs when the competing models $D$ and $G$ fail to reach Nash Equilibrium \cite{kodali2017ganconvergence} (where $p_g \approx p_r$. This may lead to issues like vanishing gradients, mode collapse and $D$ overfitting to the training data. For LDI and time series imputation, the imbalance between the $G$ and $D$ would lead to unrealistic or irrelevant imputation estimates. Several solutions proposed for LDI tasks include (1) 
adding instance noise to $D$'s inputs allows for completely continuous distributions, improving GAN stability and alleviating vanishing gradients \cite{mescheder2018ganconvergence}, 2) 
adding gradient penalty, such as in WGANs \cite{gulrajani2017wgangp}, which regularise $D$ by constraining gradient behaviour, thereby preventing mode collapse and non-convergence \cite{roth2017regularisegans}, 
3) 
balancing $D$ and $G$ by introducing autoencoder (AE) for $D$ and minimising AE losses, helping $D$ avoid overpowering $G$ 
and allowing smoother convergence \cite{bethelot2017began}, and (4) 
employing margin adaptation \cite{wang2017magan}, which regularises $D$ to balance its learning with $G$.

\textbf{Exploding bias}
problem \cite{bengio2015biasexploding} is commonly encountered in next-step prediction-based imputation with Recurrent Neural Networks (RNNs) and its variants, Long Short-Term Memory networks (LSTM) and Gated Recurrent Units (GRU) (details in section \ref{sec:ldc_gans_rec_models}). Errors in missing value estimates are often delayed until the next observed value; this increases bias and reduces training efficiency 
\cite{cao2018brits}. This can be alleviated 1) by using bi-directional RNNs (Bi-RNN) where both past and future values are used for impute estimates  \cite{gutpa2021bigan}, 2) using teacher-forcing \cite{Williams1989teacherforcing} where real data is fed into the RNN (see Figure \ref{fig:unibirnn}) so that imputation estimates at each time step are validated by observed values during training \cite{zhang2021e2ganrf}, and 3) attaching a classification objective to $G$ \cite{ma2020ajrnn}.

\section{GANs for LDI: Approaches}
\label{sec:ldi_gan_approaches}
\begin{figure}[!ht]
    \centering
    \includegraphics[width=1.0\linewidth]{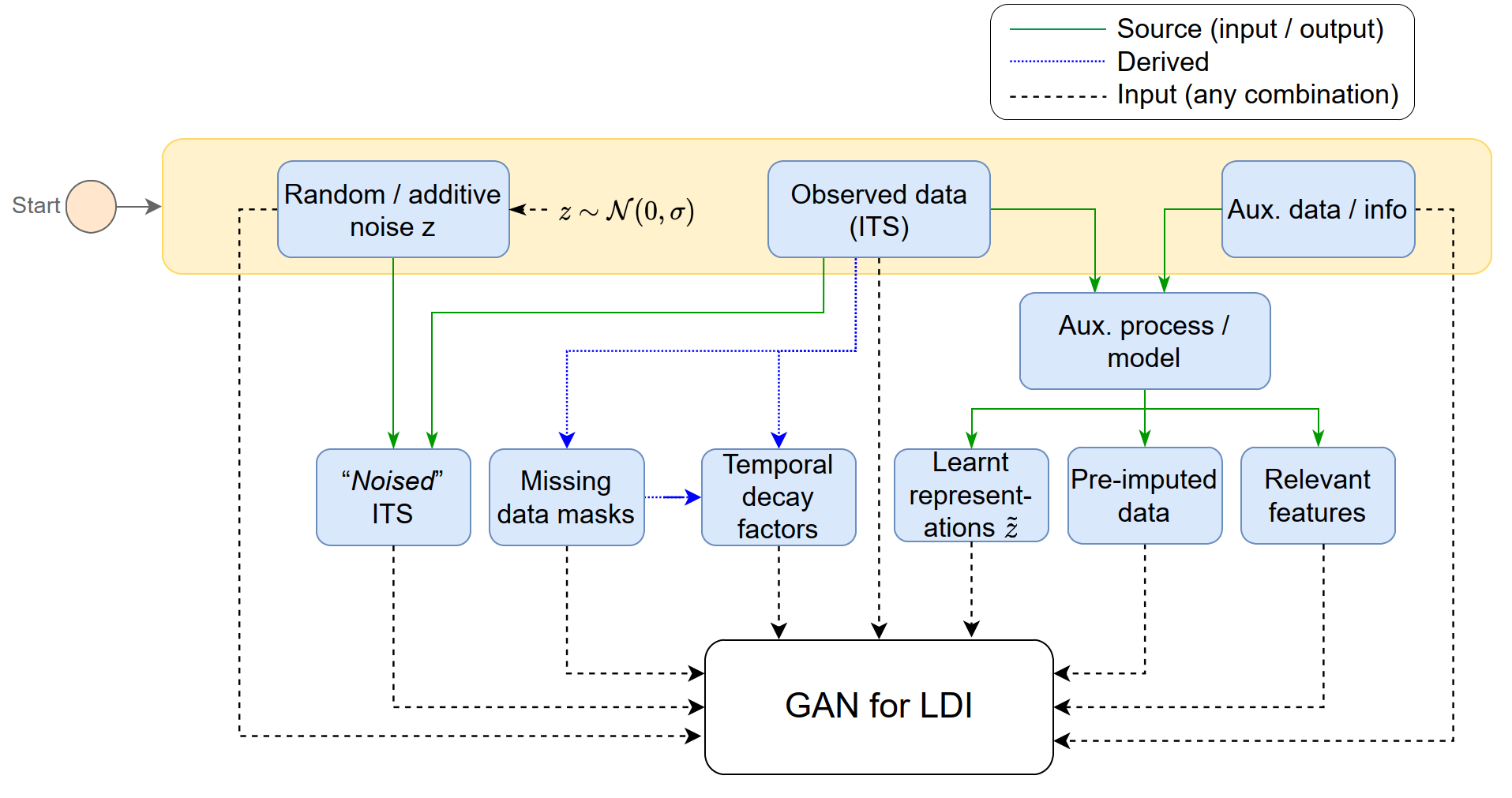}
    \caption{Inputs to GANs for LDI. Often, it is the $G$ that takes in the various inputs, depending on its requirements. \textit{From top down}: GANs may take in directly (along with other inputs) the ITS, random priors, or additional data. \enquote{Noised} ITS are from ($X+z$); masks ($M$ from Eq. \ref{eqn:missmask}) and temporal decays ($\Delta$ from Eq. \ref{eqn:timelag}) are derived from the ITS. Learnt representations $\tilde{z}=f(X)$, pre-imputed CTS (from $M\odot X + (1-M) \odot f(X)$), and relevant features ($X$ with $V$ reduced) are outputs where $f^*()$ is an auxiliary model or process.
    }
    \label{fig:g_inputs}
\end{figure}

In this section, we categorise GAN-based approaches for 
LDI depending on how and where imputation is performed in the model. In most GAN-based LDI methods, $G$ is responsible for learning the temporal dynamics of the data to facilitate imputation. The way the input data is represented and fed into GAN-based LDI plays a critical role, as shown in Figure \ref{fig:g_inputs}. Our discussion mainly focuses on the design of the $G$ and its role in imputation, highlighting the strengths and weaknesses of each approach. 

\begin{figure}[!ht]
    \centering
    \includegraphics[width=1.0\textwidth]{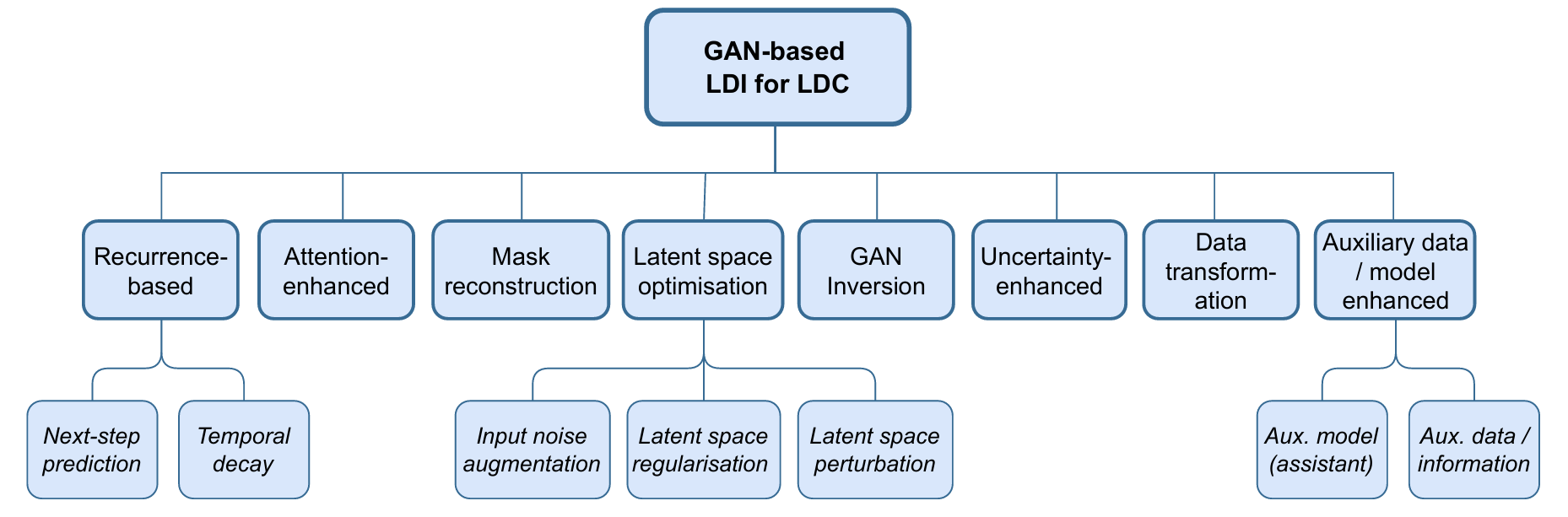}
    \caption{A taxonomy of common GAN-based Longitudinal Data Imputation (LDI) approaches for handling missing data.}
    \label{fig:imputationgans}
\end{figure}

A taxonomy of common GAN-based approaches for LDI is presented in Figure \ref{fig:imputationgans}. These methods can be distinguished based on several criteria: 1) the basic architecture, such as GAIN-based or hybrid models incorporating autoencoders, 2) the approach for the LDI task, 3) the type of inputs as listed in Figure \ref{fig:g_inputs} and how they are fed into the network, and 4) the $G$'s outputs, specifically whether $D$ assigns real/fake probabilities for the entire sample, or individual elements within it \cite{yoon2018gain}. 
Despite the dichotomy of approaches, many works combine them to leverage their respective strengths, improving imputation quality and accuracy. In this taxonomy, models are grouped according to their most significant contribution or their primary imputation method. Training and optimisation strategies are crucial for determining the success of the imputation process. Often, imputation is optimised based on the \enquote{match-worthiness} of the generated sample, either as a whole or with reference to individual elements within it. 

\subsection{Recurrence-based approach} 
\label{sec:ldc_gans_rec_models} 
This approach leverages RNN's inherent sequence-to-sequence (seq2seq) learning capability to model temporal relationships in data. These methods often employ two strategies: 1) next-step prediction, which estimates missing data during seq2seq learning by filling in the most likely values \cite{cao2018brits}, and 2) temporal decays \cite{che2017grud} as outlined in Section \ref{sec:bg_ldc_defns} (Definition 2). $G$'s final output is a fully synthetic CTS, which may either be directly passed to $D$ for adversarial training and later imputed for testing \cite{luo2018gruigan,zhang2021e2ganrf}, or imputed first and then fed to $D$ for iterative refinement of imputation estimates during training \cite{yang2023advrnnimpute}. These approaches primarily rely on recurrence models such as RNNs, GRUs and LSTMs. 

\begin{figure}[!ht]
    \centering
    \includegraphics[width=0.6\textwidth]{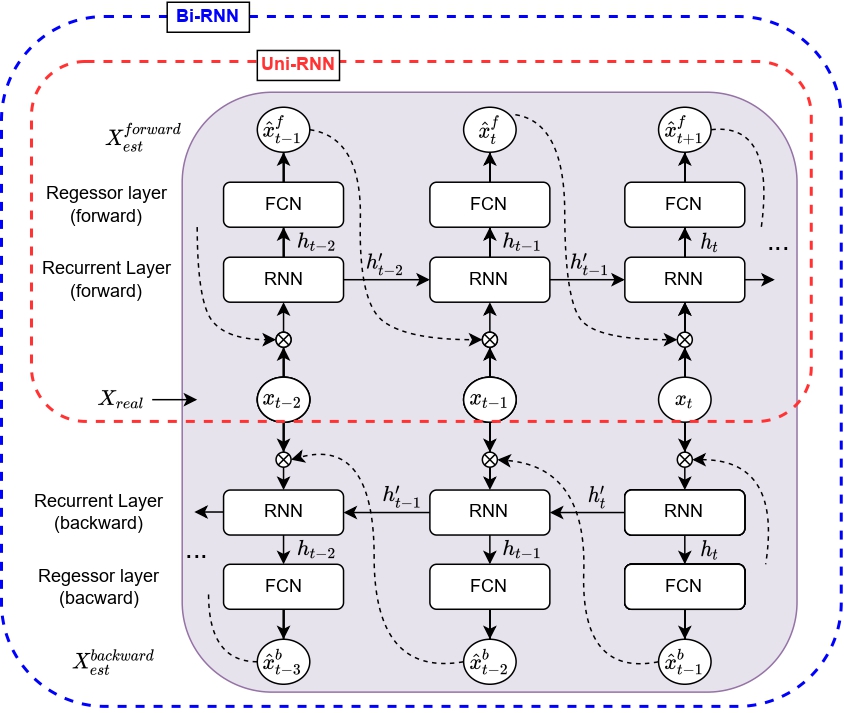}
    \caption{The self-feeding uni/bi-directional RNN cell with regression layers for predicting output of next time step which is fed back into the network using recurrent layers.}
    \label{fig:unibirnn}
\end{figure}

\noindent \textbf{Next-step prediction}, introduced by BRITS \cite{cao2018brits}, involves a uni- or bi-directional recurrent model that performs imputation of missing data by coupling a prediction layer (i.e., a linear regression layer) to a recurrent layer. This predicts the next time step's value based on the previous hidden state and temporal decay (see Figure \ref{fig:unibirnn}). GRUI-based applications, such as E\textsuperscript{2}GAN-RF \cite{zhang2021e2ganrf}, implement this as real-data forcing (RF) \cite{lamb2016professorforcing}, with the following equations: 
\begin{equation} 
    \hat{x}_{t} = W_xh_{t-1}+b_x \label{eqn:grui_regressor}
\end{equation}
\begin{equation} 
    x^c_{t} = m_t \odot x_t + (1-m_t) \odot \hat{x}_t \label{eqn:grui_impute}
\end{equation}
\begin{equation} 
    h'_{t-1}={\gamma }_{t}\odot {h}_{t-1} \label{eqn:grui_decay}
\end{equation}
where $W_x$ and $b_x$ represent the parameters of the regressor (Eq. \ref{eqn:grui_regressor}), estimating $\hat{x}_t$ from the previous hidden state $h_{t-1}$. A \textit{complementary} vector $x_t^c$ is derived using mask $m_t$ in Eqn. (\ref{eqn:missmask}) to replace estimates $\hat{x}_t$ with observed values $x_t$; improving imputation validation. This treatment allows for targeted updates during backpropagation. With a few exceptions, almost all RNN-based GANs for LDI follow a next-step prediction method, with variations in their update functions.The GRUI cell update with complementary data is given by:
\begin{equation} 
    {r}_{t}=\sigma ({W}_{r}[h'_{t-1},x^c_{t}]+{b}_{r}), \label{eqn:grui_reset}
\end{equation}
\begin{equation} 
    {z}_{t}=\sigma ({W}_{z}[h'_{t-1},x^c_{t}]+{b}_{z}) \label{eqn:grui_update}
\end{equation}
\begin{equation} 
    {\tilde{h}}_{t}=\tanh (W[{r}_{t}\odot h'_{t-1},x^c_{t}]+b_{\tilde{h}}) \label{eqn:grui_cand_ht} 
\end{equation}
\begin{equation} 
    {h}_{t}=(1-{z}_{t})\odot h'_{t-1}+{z}_{t}\odot {\tilde{h}}_{t} \label{eqn:grui_ht}
\end{equation}

Figure \ref{fig:unibirnn} shows a Uni/Bi-RNN with next step prediction. The $\otimes$ operation performed by Eqn. (\ref{eqn:grui_impute}) updates missing data only with predictions $\hat{x}_t^*$, while real values $x_t$ remain unchanged, helping refine future 
 estimates $\hat{x}_{t+1}$ using decayed hidden states $h'_{t-1}$.  

For Bi-directional RNN (BiRNN)-based $G$s in GANs, forward and backward estimates are combined, either as a mean \cite{cao2018brits,miao2021ssgan}, or a weighted sum accounting for temporal decay \cite{gutpa2021bigan}, as in Eq. (\ref{eqn:grui_impute}). The latter has shown better performance than simple averaging.
BiRNN-based GANs \cite{gutpa2021bigan,miao2021ssgan} often apply a consistency loss for forward and backward predictions \cite{cao2018brits}. Bi-GAN \cite{gutpa2021bigan}, for instance, 
combines forward and backward temporal decays for uneven sampling, focusing on 
intra-series temporal correlations. 
However, it ignores inter-series correlations, which are important for multi-variate series data \cite{sun2020irregular}. 
Although Bi-directional GRUI (BiGRUI) models have been proposed \cite{fan2021windgrui,gutpa2021bigan} for backward learning of the temporal relationships, 
most are uniGRUI-based GANs for time series imputation \cite{zhang2021e2ganrf,bi2022geda,song2023dagan} that rely on uni-directional recurrence and benefit only from past observations. 

The End-to-End GAN (E$^2$GAN) \cite{luo2019e2gan} seeks to improve the two-staged GRUI-GAN \cite{luo2018gruigan} by dynamically improving imputation quality during training. However, as shown in Figure \ref{fig:unibirnn}, RNNs are self-feeding, meaning each time step's output is fed back into the model iteratively. This risks error propagation and biased estimates \cite{ma2020ajrnn}, which the E\textsuperscript{2}GAN-RF \cite{zhang2021e2ganrf} mitigates by reducing reliance on predictions during inference using real-data forcing. 
The use of BiRNNs in models such as \cite{zhang2023imputepredgan,wu2022imgan} can reduce the gap between training and inference, reduce the impact of compound errors, and alleviate exploding biases by incorporating past and future observations, shortening the prediction path. 

The SolarGAN \cite{zhang2020solargan} also applies GRUI with noise-interpolated samples ($x+z$) before passing to $G$, allowing $G$ to capture inter-dependencies within the multivariate series. Adversarial Joint-learning RNN (AJ-RNN) \cite{ma2020ajrnn}, a uniRNN-based model, simultaneously imputes and classifies regularly sampled time series,  mitigating bias explosion and improving imputation accuracy by conditioning estimates on class labels. These methods optimise element-wise predictions and would benefit from applying 
\enquote{hints} (Eqn. (\ref{eqn:hint})) 
to ensure their $G$s learns the true distribution \cite{yoon2018gain}. The Adversarial Imputation GAN \cite{yang2023advrnnimpute} 
and semi-supervised GAN (SSGAN) \cite{miao2021ssgan} however enhance imputation quality through hinting techniques 
and classification objectives, aiding $G$ in learning a more accurate distribution conditioned on both observed values and class labels.

\subsection{Attention-enhanced}
\label{sec:ldi_gans_att}
These models use attention mechanisms \cite{Vaswani2017attention} to enhance temporal learning in RNNs and Convolutional Neural Networks (CNNs) or on their own. They aim to address challenges posed by sampling irregularity, longer sequences, sparsity and inter-feature correlations. Attention mechanisms allow the network to focus on the parts of the input data most relevant to the primary task, be it imputation or classification task.

Self-attention is the most common for time series imputation among the various attention mechanisms. It excels at learning intricate relationships within long time series without the vanishing gradient problems often encountered by RNN. In MTS, self-attention layers process a [$T \times V$] matrix, enabling the model to learn intra- and inter-feature correlations in parallel. Methods applying self-attention to GANs for imputation include Self-attention Imputation Networks using GAN (STING) \cite{oh2021sting} which uses a modified BiGRU with self and temporal attention layers; its two $G$s perform forward and backward predictions with temporal decays. Similarly, the Self-Attention Autoencoder GAN (SA-AEGAN) \cite{zhao2024saaegan} uses self-attention within a modified GRU with temporal Decay (GRU-D) \cite{che2017grud} based autoencoder (AE) to extract temporal patterns and improve imputation accuracy. GRU-D naturally assumes informative missingness \cite{che2017grud}, however SA-AEGAN employs a MCAR training regime without specifying the assumed missing data mechanism.  

Models with more complex forms of self-attention include the Pre-imputation, Parallel Convolution, and Transformer Encoder for Time Series GAN (PPCTE-TSGAIN) \cite{ma2023ppcte-tsgain}, and the Multi-head self-attention and Bidirectional RNN-based GAN (MBGAN) \cite{Ni2022mbgan}, which introduces a \enquote{feature attention} mechanism, combining multi-head self-attention \cite{saravana2024temporalgan} and BiRNNs for time series imputation. The MBGAN uses an auxiliary network to enhance performance, which is discussed further in Section \ref{sec:ldc_gans_aux_models}. 

While Transformers excel in dealing with longer sequences, they can incur significant computational overhead \cite{Vaswani2017attention}. To address this, models have utilised sparse attention, such as ProbSparse \cite{zhou2021probsparse}. 
For example, the Transformer and Generative Adversarial Network (Trans-GAN) \cite{he2023transgan}, a purely attention-based GAN, integrates the sparse attention module 
within a WGAN framework to lower computational overhead. Informer-WGAN \cite{Qian2022informergan}, which processes individual time series through multiple encoders and decoders for imputation, employs ProbSparse attention to deal with long sequences efficiently. These models aim to mitigate model collapse using WGAN losses and reduce overhead with sparse attention, however they face challenges in hyper-parameter tuning due to the complexity of their architectures. The MTS Imputation GAN (impute-GAN) \cite{qin2023imputegan} is also informer-based \cite{Zhou2021informer}. It employs an iterative strategy to improve its imputation objective, selecting only the generated samples that best match the true data for the final imputation. 

As missing values increase, local temporal dependencies become corrupted \cite{sun2020irregular}, making them less reliable \cite{tang2020lgnet}. However, attention mechanisms allow for global dependencies to be learnt which provided relevant temporal contexts for imputation. 

\subsection{Mask reconstruction approach}
\label{sec:ldi_gans_maskrecon}
GAIN \cite{yoon2018gain} (in Figure \ref{fig:gans}(f)) inspires this approach, which is useful when there is inherent data missingness. In GAIN-based models, $G$ learns a conditional distribution based on the observed data and the missing data distribution. Variables with missing values are pre-imputed or seeded with noise $z$ from a simple distribution (e.g. Gaussian) and treated as random variables, optimising for the best imputation estimates. These methods apply a reconstruction loss by comparing the imputation of observed values to their ground truth counterparts.

While GAIN is trained under the MCAR assumption, it can be extended to MAR and MNAR situations where the probability of missingness is modelled in the data generation process. This flexibility is particularly useful for longitudinal data applications, and emphasised works like the improved GAIN (iGAIN) \cite{Psychogyios2023igain} and Federated Conditional Generative Adversarial Imputation (FCGAI) \cite{zhou2020federatedcgai} models.
The iGAIN uses neighbourhood-aware AE \cite{aidos2020neighbourawareae} that take in pre-imputed ITS by KNN, using tool \cite{schouten2018generatemissing} to apply missingness patterns exhibited by MCAR, MAR and MNAR. While FCGAI applies a non-recurrent GAIN within a federated learning framework \cite{konecny2015federatedlearn}. Leveraging multiple data sources, GCGAI attempts to improve imputation quality even in fragmented or isolated datasets. 
Federated learning accounts for source heterogeneity, which is important as GANs typically assume that samples are drawn from a uniform distribution and are mutually independent. Federated learning ensures more stable training than localised GAN training, making it useful for grouped longitudinal data scenarios \cite{hassan2023fedvaelong}. 

Although the GAIN architecture is highly popular \cite{Zhang2023gainsurvey}, various adaptations have emerged to address specific challenges in LDI. For instance, to deal with complexity and resource constraints, the slim GAIN (SGAIN) \cite{gao2022sgain_slim} proposes a compact architecture that omits the hint mechanism to increase accuracy and efficiency. The Bidirectional Stackable Recurrent GAIN (BiSR-GAIN) \cite{Li2024bisrgain} uses stackable recurrent units \cite{Turkoglu2022stackablernn} to address the vanishing gradient problem. 

When applying recurrence-based GAINs to longitudinal data, it is essential to consider the missing data mechanisms and how temporal dependencies are learned during training and inferred during testing. Examples include a modified BiLSTM model with temporal decay and feature correlations \cite{yang2023advrnnimpute}, a LSTM-based GAIN framework for early sepsis prediction \cite{Apalak2022sepsispred}, and the longitudinal early classifier GAN (LEC-GAN) \cite{pingi2024lecgan} which uses imputation to improve early classification in longitudinal data. Other GAIN-based methods \cite{miao2021ssgan, yang2023advrnnimpute} that depend on next-step prediction for imputation are discussed in Section \ref{sec:ldc_gans_rec_models}. 

\subsection{Latent space optimisation-based approach}
These methods involve compressing Incomplete Time Series (ITS) and reconstructing them back from a more structured and optimised latent distribution. Autoencoders (AE) \cite{Berahmand2024encodersurvey} are often applied, assuming decoding from a lower-dimensional manifold into the feature space helps generate plausible samples to replace the incomplete data. AEs are typically pre-trained to optimise the latent space, which is then used for sample generation and reconstruction. When applied to ITS, the reconstructed samples can constitute $G$'s fake samples, particularly where the AE replaces $G$, in models like denoising autoencoders (DAE) 
\cite{luo2019e2gan,vincent2008denoiseautoencoder}. In Bi-directional GANs (BiGANs) \cite{donahue2016bigan}, GAN inversion is applied, using an AE to map real data back to their latent encodings, stabilising and improving imputation. An important consideration in this approach is that missing data disrupts temporal correlations; thus, without considering global temporal dependencies, the reconstructed space may be less reliable \cite{sun2020irregular}.

\subsubsection{Input noise augmentation}
This technique adds artificial noise to the input data (ITS) before feeding it into the model. An AE variant, Denoising AE (DAE), is often used. The learnt representations are utilised by the AE's decoder (as $G$) for reconstruction and imputation, whether jointly or having the DAE first pre-trained. The objective of the DAE is to make the model robust by learning to ignore noise and focus on the underlying data distribution. The advantage of combining GANs with DAEs for imputation is that $G$ can utilise more informative (rather than random) priors to generate CTS that are more realistic due to the GAN's adversarial training. 

Inverse Mapping GAN (IM-GAN) \cite{wu2022imgan} employs a BiRNN-based DAE, where noise is added to the ITS. The $G$ uses decoder-based outputs to reconstruct ITS. The DAE in IM-GAN learns latent representations from corrupted ITS, which are then used for imputation, generating more robust outputs in noisy data scenarios. 
This two-step process simplifies GAN training by pre-training DAE to provide informative codes to $G$ for generating samples. However, the challenge lies in the mismatch between the latent spaces of DAE and $G$ if trained separately. 
Additionally, when applied to datasets with high levels of missingness, such as medical dataset PhysioNet \cite{goldberger2000datasetphysionet}, joint training of the DAE and GAN may not be able to resolve the limitations of incomplete and fragmented data 
 \cite{pingi2024faicgan}. 

The E\textsuperscript{2}GAN \cite{luo2019e2gan}, introduced in Section \ref{sec:ldc_gans_rec_models}, also applies a DAE-based $G$, claiming that the reconstructed CTS aligns more closely with the original data for imputation than GRUI-GAN \cite{luo2018gruigan}. The Time and Location GRU GAN (TLGRU-GAN) \cite{Wang2022tlgrugan} builds on E$^2$GAN to incorporate a GRUI that accounts for both time and location intervals. 
Additionally, GRUI-based architectures have been applied within AEs for time series reconstruction \cite{bi2022geda, song2023dagan}. Trend-aware GAN (TrendGAN) \cite{li2023trendgan}, a BiGRU-based model, 
employs a GAIN-based architecture with a DAE-based $G$ to reconstruct ITS to CTS, following similar methodologies.  

While DAEs offer a degree of regularisation to AEs in learning the underlying data structure \cite{alain2014regae}, complexity is added to the joint training for such hybrid models, particularly GANs, which already suffer with instability during training. Moreover, DAEs must balance denoising and reconstruction - if the noise level is too high, the model struggles to learn meaningful relationships within the data \cite{vincent2008daechallenges}, impacting the effectiveness of the overall imputation. 

\subsubsection{Latent space regularisation}
\label{sec:ldi_gans_ae_latentreg}
This technique involves applying constraints, or other regularisation strategies to the latent space learned by $G$ to enhance the learning of complex and sparse temporal data. For instance, VAE-based $G$s apply a KL-divergence term to impose a Gaussian constraint on the latent space. The objective is to gain a more structured 
optimised latent space to draw complete samples. Without regularization, AEs applied in $G$s risk overfitting the data \cite{alain2014regae}. 

Several models employ this strategy for longer and sparser temporal sequences. The Augmented Neural Ordinary Differential Equation-assisted GAN (ANODE-GAN) \cite{chang2023anodegan} uses a VAE-based $G$, whose latent space is augmented by neural ODEs \cite{dupont2019anode}, to learn smooth and continuous temporal dynamics in irregularly sampled time series. While effective, this introduces additional computational complexity. 
Other models, such as Glow-based VAE with WGAN (GlowImp) \cite{liu2021glowimp} based on a CNN-based Glow-VAE \cite{kingma2018glow}, uses generative flows \cite{dinh2017genflow} for realistic imputations but face similar challenges regarding computational cost and hyperparameter tuning. 

The Dual Attention-enhanced GRU-based GAN (DAGAN) \cite{song2023dagan} uses temporal and relevance attention layers, and a feature aggregation layer that integrates temporal and inter-feature correlations using a residual connections \cite{he2016resnet} to improve imputation. While the GAN, Encoder-Decoder and Autoregressive network (GEDA) \cite{bi2022geda} utilises an attention-based AE and autoregressive (AR) component to better learn sequential dependencies. 
While AR methods are trained to predict values for the next time step and learn intricate inter-feature relationships, but involve complex multi-phase training.

\subsubsection{Latent space perturbation}
These methods involve perturbing the latent space with random noise to improve generalisation when reconstructing the data from the latent space, which helps particularly for longer missing time steps. Recurrent conditional GAN (rcGAN) \cite{Festag2023rcgan}  optimises hidden state outputs from an BiRNN-based contextual encoder. The encoder learns missingness patterns using local and global attention mechanisms \cite{Bahdanau2015neuralmachtrans}. Additive noise is used to initialise its hidden states and dropout layers incorporated into $G$ for probabilistic imputation, similar to DAE-GAN \cite{Zhang2024daegan}. 

Perturbations, such as noise or slight latent space alterations, further stabilise latent representations and make them robust. This approach is advantageous when the latent space is directly used for downstream tasks like classification \cite{Gabdullin2024sae}.
The Wasserstein GAIN coupled with a VAE model (WGAIN-VAE) \cite{wang2024wgainvae} addresses continuous missing data by introducing Gaussian noise in the latent space to improve generalization for reconstructing longer missing sequences. This, in principle, is similar to having dropout layers, a form of implicit noise that creates randomness in the DAE to improve the denoising process and prevent overfitting \cite{Zhang2024daegan}. This counters GAIN-based limitations, where random noise leads to disparities between $G$'s learned and true distributions \cite{yoon2018gain}. 

In principle, this approach is similar to incorporating dropout layers as a form of implicit noise that creates randomness in an autoencoder to improve the denoising process and prevent overfitting \cite{Zhang2024daegan}. Although determining the right level of noise can be challenging. Balancing noise levels is critical, as insufficient noise fails to regularise the model, while excessive noise can disrupt representation learning, degrading imputation performance; striking this balance is essential for precise imputation.

\subsection{GAN inversion} 
\label{sec:ldc_gans_inversion}
Inverse GANs \cite{Zhu2016ganinversion} focus on learning the inverse mapping from real data to its latent representation. 
GAN inversion is not a rudimentary mathematical inversion. Optimizing the latent space improves the stability and robustness of imputation. Of particular importance is the reconstruction of noisy or sparse data, where the optimisation of the latent space is critical to capturing the intricate temporal dependencies in incomplete time series or longitudinal data. This approach has become useful for applications involving reconstruction, augmentation, and ITS data manipulation. 

The Iterative GAN for Imputation (IGANI) \cite{Kazemi2021igani} iteratively refines imputation estimates by using an invertible $G$-imputer, enabling recovery of both observed and missing values from the latent space, enhancing stability and robustness.
Graph Convolutional Network (GCN)-based inverted GAN \cite{xu2023ganinvert} models temporal irregularities in ITS using temporal decays (Eq. \ref{eqn:timelag}) factored in through an adjacent matrix. It employs a pre-trained AE to reconstruct the ITS, as similarly done in the Inverse Mapping GAN (IM-GAN) \cite{wu2022imgan}.

The Partial Bi-directional GAN (P-BiGAN) \cite{li2020pbigan} models continuous ITS by utilizing continuous CNNs to learn latent space embeddings for ITS reconstruction using a VAE-GAN model. It proposes a joint supervised objective after pre-training of the encoder to be added to the $D$ loss. While these methods provide realistic estimates for missing values in time series, optimising the latent space for complex temporal dependencies can be computationally expensive.  

The Dual Transformer-based Imputation Nets (DTIN) \cite{sun2024dtin} proposes a dual GAN for incomplete and complete time series. ITS are fed through a GAIN-based architecture for imputation estimates, whose latent space is fine-tuned using a image-based pivoting technique \cite{roich2023pivotaltune}. The CTS is fed to a GAN for real/fake distinction. While novel and effective, there exists a risk that discrepancies may exist between distributions learnt by the two GANs.

\subsection{Uncertainty enhanced approach}
\label{sec:ldi_gans_uncertainty}
These GAN-based methods use probabilistic networks, random sampling or Bayesian principles to enhance the quality and accuracy of imputation and prediction tasks. C-GAN \cite{Li2024cgan_bayesian} uses transfer learning from models trained on similar datasets to obtain initial missing data estimates, which the GAN iteratively refines. Bayesian inferencing is applied with the likelihood from the pre-trained model and the prior from $G$'s learned distribution, improving forecasting with uncertainty calculations. In contrast, Uncertainty-Aware Augmented GAIN (UAA-GAIN) \cite{Feng2023uaagain} samples from the real data multiple times and calculates an uncertainty matrix from variance scores for measuring imputation loss, with a focus on complex samples \cite{Bengio2009curriculumlearning}. The Time series GAIN (TGAIN) \cite{li2023tgain} also uses multiple imputation, however after their GAN is trained, where different sets of noise are fed into $G$ for \enquote{best fill value} from the learned distribution. 

The Uncertain GAN based on GRUD (UGAN-GRUD) \cite{yin2024ugan}, proposed particularly for high missing data rates, takes in temporal decays and missing data information, calculating a normalised reconstruction loss for each time step (uncertainty matrix) which is incorporated into next-time step prediction. Similarly, rcGAN \cite{Festag2023rcgan} integrates an attention mechanism for probabilistic imputation and forecasting. Multi-Task learning-based GAIN (MT-GAIN) \cite{xu2023mtgain} applies a prediction objective to $G$ with homoscedastic uncertainty for balancing the prediction and imputation tasks. It utilises a pre-trained LSTM-FCN \cite{karim2018lstmfcn} on complete data for prediction, assuming a sufficient and complete representative data sample for training. 

Incorporating uncertainty into GAN-based LDI or prediction tasks helps models adapt to discrepancies between generated and real data, improving imputation accuracy, particularly with high missing data rates. However, if the uncertainty information does not accurately reflect the underlying data structure, there is the risk of overfitting to noise, potentially reducing performance on unseen data with unpredictable missing patterns.

\subsection{Data transformation approach}
Methods discussed here transform ITS data into another form to address the missing data problem. Most commonly, the ITS is transformed into images, and a semantic \enquote{in-painting} approach \cite{yeh2016inpainting} is then applied for imputation. Often, the data is divided into CTS and ITS subsets, where ITS is seen as the \enquote{corrupted} version of the dataset. CTS is then used to train a GAN; however, if the ITS does not align with either of the learned manifolds $p_r$ and $p_g$, this poses a challenge. After model optimisation, the closest latent encoding for the ITS is identified using a reconstruction loss, similar to GAN inversion. These methods rely on two assumptions: (1) ground truth is available to train the GAN, which is challenging in real-world longitudinal data with irregular sampled time series, and (2) the CTS subset is representative of the whole dataset, which can be misleading.

To transform time series data into images, methods such as Gramian Angular Summation Field (GASF) \cite{Thanaraj2020gasf} are used to preserve local and global temporal dependencies. Traffic Sensor Data Imputation GAN (TSDIGAN) \cite{huang2023tsdigan} uses GASF to represent time series as images using DC-GAN \cite{radford2015dcgan}, thus shifting the task from time-series imputation to image in-painting. A maximum mean discrepancy loss function \cite{li2017mmdgan} is used to measure statistical similarities between the generated $p_g$ and real data distribution $p_r$. While the use of convolutional layers allows the model to capture spatiotemporal relationships more effectively, they risk missing finer temporal correlations.  

In some methods, temporal data is aggregated to represent various statistical features, which removes the temporal aspect of longitudinal data. To deal with non-numeric data, the Categorical EHR imputation with GAIN (Categorical GAIN) \cite{yang2019categoricalgain} proposes a fuzzy binary coding, allowing the GAIN framework to process multi-class and multi-label data from the start of training. Temporal features here are aggregated, normalised \cite{Esteban2015latentembed} and then modelled with static and other non-temporal data. While this provides a novel solution to handling categorical features, it does not consider the joint modelling of both categorical and numerical features. 

\subsection{Auxiliary data-enhanced approach}
This involves incorporating additional data sources like class labels, other similar datasets, modalities or dimensions (like static data) and separate parts of the data to exploit correlations or inter-dependencies between these sources and improve the imputation process. This approach is often beneficial when the time series data is compromised by excessive noise or missing values, as reported by the Fusion-aided Imputer/Classifier GAN (FaIC-GAN) \cite{pingi2024faicgan}. FaIC-GAN proposes several fusion strategies to integrate the static component of longitudinal datasets with temporal data for improved classification. This method assumes that the joint representation of static and temporal components provides more informative features for both the imputation and subsequent classification tasks. 

The Compressive Population Health (CPH) model \cite{feng2021cph} organises input data into classes and uses a GAIN-based model to learn distributions of both observed and missing data within each class, unlike methods \cite{awan2021imputationcgain, li2021eegcontextgan, Apalak2022sepsispred} that condition imputation on class labels. The clinical conditional GAN (ccGAN) \cite{Bernardini2023ccgan} uses fully observed time series to guide imputation. While shown to improve classification, ccGAN risks losing global temporal and missing data correlations \cite{sun2020irregular}, given its assumption of non-temporality within short and sparsely recorded clinical data and the discard of over 50\% of features with high missing rates. 

Impute-GAN \cite{qin2023imputegan}, introduced in Section \ref{sec:ldi_gans_att}, improves imputation by training self-encoders on small datasets with missing values. The improved Conditional GAN (C-GAN) \cite{Li2024cgan_bayesian} uses transfer-learning to train a separate network on similar (complete) datasets to improve forecasting on their target (incomplete) data, as further discussed in Section \ref{sec:ldi_gans_uncertainty}. While C-GAN ensures that the auxiliary data share similar geographical characteristic, individual time series display unique temporal patterns due to temporal distribution shifts that necessitate some domain adaptation to bridge distribution gaps \cite{du21adarnn}.  

\subsection{Auxiliary model-enhanced approach}
\label{sec:ldc_gans_aux_models}
These methods use auxiliary techniques to enhance model performance. While AEs and their denoising and variational variants can be included in this section, they are discussed separately due to their widespread applications with GANs for LDI. Methods that incorporate statistical or machine learning techniques for ITS imputation include the I-GAIN \cite{Psychogyios2023igain} and PPCTE-TSGAIN \cite{ma2023ppcte-tsgain} which uses KNN and MICE estimates for missing data and pass to $G$ to refine further. PPCTE-TSGAIN uses parallel convolutional and multi-head self-attention layers to capture intricate data relationships. Using KNN-generated seed values instead of random priors speeds up convergence, as the GAN starts from a more informed position. However, KNN's temporal insensitivity and reliance on distance metrics can lead to inaccurate imputation for sparse or noisy data, potentially causing biased imputation if the GAN fails to capture underlying distributions \cite{qiu2024ganpriors}. 

This GAN-based model \cite{ou2024rfgain} uses a dual interpolation strategy, using random forest and GAIN for missing data estimates. Other works \cite{li2021eegcontextgan, Apalak2022sepsispred} attach auxiliary classifiers to learn contextual information or classify imputed samples. Methods like the Classification and Clustering-based GAIN (CC-GAIN) \cite{Hwang2024ccgain} and Time-series Convolutional GAN (TCGAN) \cite{huang2023tcgan2} also integrate a classification objective into their GAN-based LDI framework. 
The Fully Connected Network (FCN)-based CC-GAIN classifies $G$'s imputed samples using pseudo-labels generated by K-means clustering; while the Time series Conditional Dual-Discriminator GAN (TsCDD-GAN) \cite{peng2024tscddgan} uses a separate cluster $D$ to assist \enquote{cluster-friendly} sample generation. 
Still, there is the risk of losing temporal relationships. TCGAN adopts a two-phased strategy, using the trained $D$'s intermediate layers for transfer learning. While this limits the GAN's potential benefit from the classification/clustering objective, it reduces complexity compared to multi-task learning in $D$s.

For joint modelling of Local and Global temporal dynamics and MTS forecasting, LGnet \cite{tang2020lgnet} leverages knowledge from an external memory module \cite{weston2010memorynetwork} fed with CTS to capture global correlations, ensuring that temporally similar sequences can be used for contextual imputation. However, similar to MT-GAIN \cite{xu2023mtgain} (discussed in Section \ref{sec:ldi_gans_uncertainty}) that uses a pretrained classifier, relying on a complete subset assumes it is representative of the entire dataset. MBGAN \cite{Ni2022mbgan} employs a multi-head self-attention-based AE for $G$, utilising recursive feature extraction via XGBoost and SVMs, along with forward and backward temporal decay calculations for each feature's reliability. While improving imputation accuracy, risk of useful correlations being lost from discarded features, deemed \enquote{irrelevant}, can potentially affect the overall model performance. 

\subsection{Other approaches}
Here, we discuss some unique approaches to GAN-based time series imputation worth mentioning. The Progressive-growing and Self-Attention Convolutional GAN (PSA-CGAN) \cite{gao2024psacgan} proposes learning coarse-to-finer ITS details to capture both short and long term temporal dependencies. PSA-GAIN utilises CNNs, sparse normalisation and progressive growth for context-free forecasting that ensures training stability. Coarser time series data is processed first, and progressively finer details are added.

The Missing Values Imputation and Imbalanced learning GAN (MVIIL-GAN) \cite{weng2024mviilgan} jointly optimises imputation and minority class sample generation. An AE-based $G$ learns low-level representations from real data and fake samples of minority classes, while duel discriminators $D$s focus on sample- and variable-level discrimination. This joint learning efficiently balances the two tasks, though fine-tuning the hyperparameters is critical for optimizing imputation and minority sample generation. 

The Stackelberg GAN \cite{zhang2018stackelbergganmed}, an early work on medical data imputation introduces an ensemble of FCN-based $G$s to diversify generated values for imputation and combat mode collapse. For each missing value, a $G$ is randomly drawn from a pool to obtain a sample for imputation. While this method improves imputation quality and mitigates mode collapse, it does not consider temporal correlations both within and across features. Though non-longitudinal, the Multivariate Time Series GAN (MTS-GAN) \cite{guo2019mtsgan} is worth mentioning for its use of multi-channelled CNNs within a DC-GAN architecture to address missing data in MTS. The Slim GAIN (SGAIN) \cite{geng2024slimgain}, introduced in Section \ref{sec:ldi_gans_maskrecon} proposes the idea of having slim $G$'s and $D$'s for accurate imputations. 

Other interesting LDI works include the DAGAN \cite{song2023dagan}, which uses a residual network to integrate representations from its temporal and relevance attention layers for more effective imputation. Model \cite{xu2023ganinvert} which uses Graph Convolutional Networks for more precise reconstruction of missing values estimates. The DTIN \cite{sun2024dtin} is considered notable for its image-style manipulation of time series data for improved imputation. 

\subsection{Summary of GANs for LDI}
This section reviewed various GAN-based approaches to LDI, with the main goal of generating samples that closely resemble real data for best imputation estimates. The most common approach to LDI is recurrence-based methods, mostly based on GRUI and Bi-directional RNNs that incorporate time decay. Since RNNs suffer from vanishing gradients, they are often temporally enhanced by attention mechanisms. CNN-based methods are also common, particularly for irregularly sampled longitudinal data, although these assume the individual time series are i.i.d.. A large number of methods employ auxiliary models, particular to learn latent representations from complete subsets of the data, although these assume that these subsets are adequate representations of the data, and that they are always available. The third most common approach is AE-based, where most methods replace with generators for reconstruction from a more structured and robust latent space. To ensure this, methods have to regulate the latent encodings or add noise to the data that needs balancing. Works often employ a mixture of approaches to leverage the strength of each approach. 

Given the focus on LDC, literature evidence shows widespread efforts to cooperate classification and/or clustering objectives, particularly joint tasks \cite{ma2020ajrnn, weng2024mviilgan, pingi2024lecgan, li2020pbigan, sun2024dtin, peng2024tscddgan} for shared representations and improved imputation estimates. To handle long, sparse, or high-dimensional ITS, some methods employ attention mechanisms \cite{Vaswani2017attention}, such as self-attention or multi-head attention. Others use auxiliary clustering or classification modules, integrate class labels, or leverage static features to improve imputation. The mask reconstruction approach, based on GAIN \cite{yoon2018gain}, is widely adopted. Other approaches focus on uncertainty or transform ITS into images, addressing LDI as an image inpainting problem. These complex models for imputation often come with higher training times, added complexity and a need for fine-tuned hyper-parameters. 

Figure \ref{fig:topolgy_challenges} presents a taxonomy of common GAN-based LDI challenges. Most methods do not explicitly state the assumed missing data mechanism, defaulting to MCAR \cite{yoon2018gain} when random masking is part of training. Certain temporal data learning structures like GRU-D \cite{che2017grud} naturally assume MNAR, this should align with model evaluation strategies. Additionally, static features are largely ignored by GAN-based LDI, with methods assuming i.i.d. for time series data related to individual instances. While some methods aim particularly for irregularly sampled data \cite{xu2023ganinvert,chang2023anodegan}, many address random missing in the data. In terms of class imbalance, only one method covered directly addressed class imbalance \cite{weng2024mviilgan}, while others resample the data \cite{pingi2024faicgan,Psychogyios2023igain}; most only acknowledged the problem and used AUC to judge the robustness of their model \cite{xu2023mtgain,li2020pbigan,liu2021glowimp}. A couple of methods \cite{pingi2024faicgan,pingi2024lecgan} addressed instance heterogeneity, though most discard these features as non-essential to the imputation or downstream task. Mixed data types in longitudinal data is mostly ignored. Although there exist many methods that treat these data-level challenges separately.
 
Tables \ref{tab:recurrence_approaches}, \ref{tab:temp_approaches} and \ref{tab:nonrecurrence_approaches} present a summary of state-of-the-art recurrence-based, Attention-enhanced methods and non-recurrence based GAN methods employed in LDI, some of which aim directly to improve the performance of LDC.

\begin{sidewaystable}[!ht]
\scriptsize 
\centering
\caption{Recurrence-based approaches in GANs for LDI}
\label{tab:recurrence_approaches}
\begin{tabular}{|ll|c|l|l|l|l|l|}
\hline
\multicolumn{2}{|c|}{\textbf{Model name}} & \textbf{Year} & \textbf{Problem addressed} & \textbf{\begin{tabular}[c]{@{}l@{}}Approaches \& \\ enhancements\end{tabular}} & \textbf{Base model (G)} & \textbf{\begin{tabular}[c]{@{}l@{}}Other challenges\end{tabular}} & \textbf{Data} \\ \hline

\cite{luo2019e2gan} & E2GAN & 2019 & Two-stage GAN issues & z-optimise & GRU-I, WGAN & 
\begin{tabular}[c]{@{}l@{}}Imbalanced \\ (AUC)\end{tabular} & 
\begin{tabular}[c]{@{}l@{}}PhysioNet 2012, \\ KDD 2018\end{tabular} \\ \hline

\cite{zhang2020solargan} & SolarGAN & 2020 & Solar forecasting & Recurrence (Alone) & GRU-I, WGAN & MAR & Solar \\ \hline

\cite{ma2020ajrnn} & AJ-RNN & 2020 & 
\begin{tabular}[c]{@{}l@{}}Classification + exploding \\ bias\end{tabular} & Aux. model (Class.) & GRU-I, cGAN & MCAR & 
\begin{tabular}[c]{@{}l@{}}4x UCR Incomplete \\ time series\end{tabular} \\ \hline

\cite{tang2020lgnet} & LGNet & 2020 & 
\begin{tabular}[c]{@{}l@{}}MTS forecasting, Local \& \\ global temp. dependencies\end{tabular} & 
\begin{tabular}[c]{@{}l@{}}Aux. model (memory \\ module for global \\ correlations)\end{tabular} & LSTM, GAN & - & 
\begin{tabular}[c]{@{}l@{}}PhysioNet 2012, \\ KDD 2018, Taxi, Weather\end{tabular} \\ \hline

\cite{zhang2021e2ganrf} & E2GAN-RF & 2021 & Two-stage GAN issues & Recurrence (Alone) & GRU-I, WGAN & 
\begin{tabular}[c]{@{}l@{}}Imbalanced \\ (AUC)\end{tabular} & 
\begin{tabular}[c]{@{}l@{}}PhysioNet 2012, \\ KDD 2018\end{tabular} \\ \hline

\cite{miao2021ssgan} & SSGAN & 2021 & Unlabelled data & 
\begin{tabular}[c]{@{}l@{}}Mask recon. + Aux. \\ model (Classifier)\end{tabular} & GRU-I, cGAIN & 
\begin{tabular}[c]{@{}l@{}}Imbalanced (AUC), \\ MCAR\end{tabular} & 
\begin{tabular}[c]{@{}l@{}}PhysioNet 2012, \\ KDD 2018, Activity\end{tabular} \\ \hline

\cite{gutpa2021bigan} & Bi-GAN & 2021 & 
\begin{tabular}[c]{@{}l@{}}Irregular sampling \& \\ dropout EHR\end{tabular} & Recurrence (Alone) & 
\begin{tabular}[c]{@{}l@{}}BiRNN (GRU-I-like), \\GAN \end{tabular} & - & 
\begin{tabular}[c]{@{}l@{}}2x EHR (Nemours \\ Pediatric, All of us)\end{tabular} \\ \hline

\cite{fan2021windgrui} & - & 2021 & Wind speed analysis & Recurrence (Alone) & Bi-GRU-I, WGAN & Continuous MVs & Wind power \\ \hline

\cite{Apalak2022sepsispred} & - & 2022 & 
\begin{tabular}[c]{@{}l@{}}Early prediction of \\ Sepsis\end{tabular} & 
\begin{tabular}[c]{@{}l@{}}Mask recon. + Aux. \\ data \& model (Class.)\end{tabular} & LSTM, GAIN & Weighted CE & 
\begin{tabular}[c]{@{}l@{}}PhysioNet 2019\\ ICU\end{tabular} \\ \hline

\cite{bi2022geda} & GEDA & 2022 & 
\begin{tabular}[c]{@{}l@{}}Water   monitor equip.\\      data for tasks like \\      classification\end{tabular} & 
\begin{tabular}[c]{@{}l@{}}z-optimise (auto-\\regressive) \end{tabular} & GRU-I, WGAN & AUC & Water   quality \\ \hline
 
\cite{wu2022imgan} & IM-GAN & 2022 & 
 \begin{tabular}[c]{@{}l@{}}Indoor   air quality \\      analysis\end{tabular} & 
 \begin{tabular}[c]{@{}l@{}}z-optimise   + GAN \\      invert\end{tabular} & 
 \begin{tabular}[c]{@{}l@{}}BiRNN   + Time \\      decay, WGAN\end{tabular} & - & 2x   Air quality \\ \hline
 
\cite{Wang2022tlgrugan} & 
 \begin{tabular}[c]{@{}l@{}}TLGRU-\\      GAN\end{tabular} & 2022 & MVs   in MTS & z-optimise & TLGRU, WGAN & - & 
 \begin{tabular}[c]{@{}l@{}}KDD   2018, \\      Air quality, Weather\end{tabular} \\ \hline
 
\cite{yang2023advrnnimpute} & - & 2023 & 
 \begin{tabular}[c]{@{}l@{}}Informative   dependencies, \\ correlations over time.\end{tabular} & Mask   recon. & GRU-I, GAIN & 
 \begin{tabular}[c]{@{}l@{}}Imbalanced   \\      (AUC)\end{tabular} & 
 \begin{tabular}[c]{@{}l@{}}PhysioNet   2012, Air \\      quality, Meteorological\end{tabular} \\ \hline
 
\cite{zhang2023imputepredgan} & - * & 2023 & Smart   grid monitoring & Mask   recon. & 
 \begin{tabular}[c]{@{}l@{}}BiLSTM   + Time \\      decay + RF, GAIN \end{tabular} & - & Gas   concentration \\ \hline
 
\cite{chang2023anodegan} & 
 \begin{tabular}[c]{@{}l@{}}ANODE-\\      GAN\end{tabular} & 2023 & 
 \begin{tabular}[c]{@{}l@{}}Imputation   at desired \\  time points, preserve \\ temporal dynamics.\end{tabular} & 
 \begin{tabular}[c]{@{}l@{}}z-optimise   + Aux. \\      model (ANODE)\end{tabular} & 
 \begin{tabular}[c]{@{}l@{}}Time-aware   \\      LSTM, cGAN\end{tabular} & MCAR & 
 \begin{tabular}[c]{@{}l@{}}Gas   sensor, \\      Air quality, \\      Electricity\end{tabular} \\ \hline
 
\cite{li2023trendgan} & 
 \begin{tabular}[c]{@{}l@{}}Trend-\\      GAN\end{tabular} & 2023 & 
 \begin{tabular}[c]{@{}l@{}}Implied   trends in MTS, \\      Continuous missing\end{tabular} & 
 \begin{tabular}[c]{@{}l@{}}z-optimise   (latent \\      regularise)\end{tabular} & BiGRU, GAN & Not   MCAR & 
 \begin{tabular}[c]{@{}l@{}}Electricity,   Weather, \\  Spam, Water quality\end{tabular} \\ \hline
 
 \cite{li2023tgain} & TGAIN & 2023 & 
 \begin{tabular}[c]{@{}l@{}}Multistate   in time series, \\      imputation uncertainty\end{tabular} & 
 \begin{tabular}[c]{@{}l@{}}z-optimise   + \\      Uncertainty\end{tabular} & LSTM + FCN, GAIN & MCAR,   MAR & 2x   Traffic \\ \hline
 
\cite{wang2024wgainvae} & 
 \begin{tabular}[c]{@{}l@{}}WGAIN-\\      VAE\end{tabular} & 2024 & 
 \begin{tabular}[c]{@{}l@{}}Imputation   for \\      continuous MVs\end{tabular} & z-optimise & MLP, WGAIN & 
 \begin{tabular}[c]{@{}l@{}}Continuous   \\      MVs\end{tabular} & 
 \begin{tabular}[c]{@{}l@{}}KDD   2018, Stock, \\      Transformers\end{tabular} \\ \hline
 
\cite{peng2024tscddgan} & 
 \begin{tabular}[c]{@{}l@{}}TsCDD-\\      GAN*\end{tabular} & 2024 & 
 \begin{tabular}[c]{@{}l@{}}Unified   framework for \\      clustering \& imputation\end{tabular} & 
 \begin{tabular}[c]{@{}l@{}}z-optimise   + Aux. \\      model (Clustering)\end{tabular} & Bi-RNN, cGAN & - & 20x   UCR datasets \\ \hline
 
\cite{Li2024bisrgain} & 
 \begin{tabular}[c]{@{}l@{}}BiSR-\\      GAIN\end{tabular} & 2024 & 
 \begin{tabular}[c]{@{}l@{}}Specific   emitter \\      identification\end{tabular} & Mask   recon. & BSR   units, GAIN & 
 \begin{tabular}[c]{@{}l@{}}Continuous   \\      MVs\end{tabular} & Emitter   signals \\ \hline
 
\cite{pingi2024lecgan} & 
 \begin{tabular}[c]{@{}l@{}}LEC-\\      GAN\end{tabular} & 2024 & 
 \begin{tabular}[c]{@{}l@{}}Early   classification  \\      with MV\end{tabular} & 
 \begin{tabular}[c]{@{}l@{}}Aux. data \& model \\      (Class.)\end{tabular} & LSTM, cGAIN & 
 \begin{tabular}[c]{@{}l@{}}Imbalanced   \\      SMOTE, \\      MCAR, Static\end{tabular} & 
 \begin{tabular}[c]{@{}l@{}}PhysioNet,   Education, \\      OASIS\end{tabular} \\ \hline
 
\cite{yin2024ugan} & 
\begin{tabular}[c]{@{}l@{}}UGAN-\\      GRUD\end{tabular} & 2024 & 
\begin{tabular}[c]{@{}l@{}}Biomed.   data analysis; \\ high missing rates\end{tabular} & Uncertainty & GRU-D, GAN & - & 4x   Health data \\ \hline
\end{tabular}%
\footnotetext{\textbf{Abbreviations}: z-opimise \textit{- latent space optimisation approach;} Aux. \textit{- Auxiliary;} Rec. - \textit{recurrence-based approach;} Att. \textit{- Attention-based approach}\textit{;} recon. \textit{- reconstruction; }MV \textit{- missing values.}}
\end{sidewaystable}

\begin{sidewaystable}[!ht]
\scriptsize 
\centering
\caption{Attention-enhanced approaches in GANs for LDI}
\label{tab:temp_approaches}
\begin{tabular}{|ll|c|l|l|l|l|l|}
\hline
\multicolumn{2}{|c|}{\textbf{Model name}} & \textbf{Year} & \textbf{Problem addressed} & \textbf{\begin{tabular}[c]{@{}l@{}}Approaches \& \\ enhancements\end{tabular}} & \textbf{Base model (G)} & \textbf{\begin{tabular}[c]{@{}l@{}}Other challenges\end{tabular}} & \textbf{Data} \\ \hline

\cite{oh2021sting} & STING & 2021 & 
\begin{tabular}[c]{@{}l@{}}MTS   correlations, \\      inherent MV\end{tabular} & Rec. + Att.& 
\begin{tabular}[c]{@{}l@{}}GRU   + Att.(Self \& \\      Temporal), WGAN\end{tabular} & MCAR & 
\begin{tabular}[c]{@{}l@{}}PhysioNet   2012, \\      KDD 2018, \\      Gas sensor\end{tabular} \\ \hline

 \cite{Ni2022mbgan} & MBGAN & 2022 & 
 \begin{tabular}[c]{@{}l@{}}Correlation   in long \\      sequences, relevant \\      features in MTS\end{tabular} & 
 \begin{tabular}[c]{@{}l@{}}Rec. + Att.+ Aux. \\model (Feature extract)\end{tabular} & 
 \begin{tabular}[c]{@{}l@{}}BiGRU   + Att.\\      (Multi-head), GAIN\end{tabular} & MCAR & 
 \begin{tabular}[c]{@{}l@{}}MIMIC-III,   Air quality, \\      water quality\end{tabular} \\ \hline

\cite{Qian2022informergan} & 
\begin{tabular}[c]{@{}l@{}}Informer-\\      WGAN\end{tabular} & 2022 & High   missing rates & Att.+ z-optimise & 
\begin{tabular}[c]{@{}l@{}}CNN   + Att.\\ (Sparse), WGAN\end{tabular} & - & 
\begin{tabular}[c]{@{}l@{}}Electricity,   Power \\      load, Telemetry TS\end{tabular} \\ \hline

 \cite{Festag2023rcgan} & rcGAN & 2023 & 
 \begin{tabular}[c]{@{}l@{}}Forecasting   \& \\      uncertainty with \\      imputation\end{tabular} & 
 \begin{tabular}[c]{@{}l@{}}Rec. + Att.+ z-optimise  \\       + Uncertainty\end{tabular} & \begin{tabular}[c]{@{}l@{}}BiLSTM  + RF + \\Att, WGAN\end{tabular} & - & 
 \begin{tabular}[c]{@{}l@{}}PhysioNet   Automatic \\      Aging\end{tabular} \\ \hline
 
 \cite{song2023dagan} & DAGAN & 2023 & MVs   in MTS & 
 \begin{tabular}[c]{@{}l@{}}z-optimise.   + Att.(with \\      Residual conn.)\end{tabular} &\begin{tabular}[c]{@{}l@{}} GRU-I   + Att, \\cGAIN\end{tabular} & - & 
 \begin{tabular}[c]{@{}l@{}}Meteorological,   \\      Health-care\end{tabular} \\ \hline

 \cite{ma2023ppcte-tsgain} & 
 \begin{tabular}[c]{@{}l@{}}PPCTE-\\      TSGAIN*\end{tabular} & 2023 & MVs   in MTS & 
 \begin{tabular}[c]{@{}l@{}}Att. + z-optimise   + \\Mask  recon + Aux. \\Model      (Pre-imputation)\end{tabular} & 
 \begin{tabular}[c]{@{}l@{}}Parallel   CNN + \\ Att, GAIN \end{tabular} & - & 
 \begin{tabular}[c]{@{}l@{}}Aeroengine   healh \\      (simulation)\end{tabular} \\ \hline
 
\cite{he2023transgan} & 
\begin{tabular}[c]{@{}l@{}}Trans-\\      GAN\end{tabular} & 2023 & 
\begin{tabular}[c]{@{}l@{}}MVs   in MTS,  inadequate \\      reasoning under high MV\end{tabular} & Att. + z-optimise & Att.   (Sparse), WGAN & - & \begin{tabular}[c]{@{}l@{}}Electricity,   enery, \\      weather\end{tabular} \\ \hline

\cite{xu2023ganinvert} & Xu2023 & 2023 & 
\begin{tabular}[c]{@{}l@{}}Optimal   latent codes for \\      ITS for precise \\      reconstruction\end{tabular} & 
\begin{tabular}[c]{@{}l@{}}Att. + GCN   + Aux. \\ model  (Latent encodings) \\      + GAN Invert\end{tabular} & 
\begin{tabular}[c]{@{}l@{}}Graph Conv. Network \\  (GCN), GAN \end{tabular} & MCAR & 
\begin{tabular}[c]{@{}l@{}}Sine,   Stock, Energy, \\      Air quality, Letter, \\      Spam\end{tabular} \\ \hline
 
 \cite{qin2023imputegan} & 
 \begin{tabular}[c]{@{}l@{}}impute-\\      GAN\end{tabular} & 2023 & 
 \begin{tabular}[c]{@{}l@{}}MVs   in MTS, unstable \\      imputation results\end{tabular} & 
 \begin{tabular}[c]{@{}l@{}}Att. + z-optimise + \\ Aux. data      (small CTS)\end{tabular} & Att.   (Sparse), GAN & 
 \begin{tabular}[c]{@{}l@{}}Continuous   \\      MVs\end{tabular} & 
 \begin{tabular}[c]{@{}l@{}}2x   Electricity, \\      Weather, \\      KDD Weather\end{tabular} \\ \hline
 
 \cite{zhao2024saaegan} & 
 \begin{tabular}[c]{@{}l@{}}SA-\\      AEGAN\end{tabular} & 2024 & QAR   analysis & Rec. + Att. + z-optimise & GRU-D   + Att., GAN & MNAR   (method) & \begin{tabular}[c]{@{}l@{}}PhysioNet   2012,\\      QAR records\end{tabular} \\ \hline

 \cite{pingi2024faicgan} & 
 \begin{tabular}[c]{@{}l@{}}FaIC-\\      GAN\end{tabular} & 2024 & 
 \begin{tabular}[c]{@{}l@{}}Classification   with MV \& \\      instance heterogeneity\end{tabular} & 
 \begin{tabular}[c]{@{}l@{}}Rec. + Att. + Mask  \\ recon +  Aux.    data \\ \& model (Class.)\end{tabular} & LSTM + Att., cGAIN & 
 \begin{tabular}[c]{@{}l@{}}Imbalanced   \\      SMOTE, \\      MCAR, Static\end{tabular} & 
 \begin{tabular}[c]{@{}l@{}}MIMIC-III,   PhysioNet, \\      Education, OASIS\end{tabular} \\ \hline
 
 \cite{gao2024psacgan} & 
 \begin{tabular}[c]{@{}l@{}}PSA-\\      CGAN\end{tabular} & 2024 & 
 \begin{tabular}[c]{@{}l@{}}Context-free   focasting \\      (bridge performance), \\      long-term dependencies\end{tabular} & 
 \begin{tabular}[c]{@{}l@{}}Other (Progressively  \\      growing GAN)\end{tabular} & \begin{tabular}[c]{@{}l@{}}CNN   + Att. (Self), \\cGAN\end{tabular} & - & Railway   bridges \\ \hline
 
 \cite{saravana2024temporalgan} & - & 2024 & MVs   in MTS & Att. + z-optimise & \begin{tabular}[c]{@{}l@{}}Att.   (Multi-head),\\ GAN \end{tabular} & MCAR & 
 \begin{tabular}[c]{@{}l@{}}Air   quality, \\      Electricity\end{tabular} \\ \hline
 
 \cite{sun2024dtin} & DTIN & 2024 & 
 \begin{tabular}[c]{@{}l@{}}Specific   emitter \\      identification\end{tabular} & 
 \begin{tabular}[c]{@{}l@{}}Att. + Mask   recon. + \\ Aux. model + GAN \\ invert (+ pivotal tuning)\end{tabular} & \begin{tabular}[c]{@{}l@{}}Att.   + 1D CNN, \\GAN \end{tabular} & - & \begin{tabular}[c]{@{}l@{}}2x   Emitter signals \\      (Airplane, Bluetooth)\end{tabular} \\ \hline

\end{tabular}%
\footnotetext{\textbf{Abbreviations}: z-opimise \textit{- latent space optimisation approach;} Aux. \textit{- Auxiliary;} Rec. - \textit{recurrence-based approach;} Att. \textit{- Attention-based approach}\textit{;} recon. \textit{- reconstruction; }MV \textit{- missing values.}}
\end{sidewaystable}

\begin{sidewaystable}[!ht]
\scriptsize 
\centering
\caption{Non-recurrence based approaches in GANs for LDI}
\label{tab:nonrecurrence_approaches}
\begin{tabular}{|ll|c|l|l|l|l|l|}
\hline
\multicolumn{2}{|c|}{\textbf{Model name}} & \textbf{Year} & \textbf{Problem addressed} & \textbf{\begin{tabular}[c]{@{}l@{}}Approaches \& \\ enhancements\end{tabular}} & \textbf{Base model (G)} & \textbf{\begin{tabular}[c]{@{}l@{}}Other challenges\end{tabular}} & \textbf{Data} \\ \hline

\cite{guo2019mtsgan} & \begin{tabular}[c]{@{}l@{}}MTS-\\      GAN*\end{tabular} & 2019 & MVs   in MTS & \begin{tabular}[c]{@{}l@{}} Other (conv./deconv.\\ GAN, i.e. DCGAN) \end{tabular} & \begin{tabular}[c]{@{}l@{}}Multi-channel   \\      CNN, DCGAN\end{tabular} & - & \begin{tabular}[c]{@{}l@{}}Synthetic   x2, machine \\      fault detection\end{tabular} \\ \hline
 
 \cite{yang2019categoricalgain} & 
 \begin{tabular}[c]{@{}l@{}}categorical-\\       GAIN*\end{tabular} & 2019 & 
 \begin{tabular}[c]{@{}l@{}}Categorical   EHR data \\      representation\end{tabular} & 
 \begin{tabular}[c]{@{}l@{}}Mask   recon. + \\      Data transform\end{tabular} & 
FCN, GAIN& 
 \begin{tabular}[c]{@{}l@{}}Categorical   \\      data\end{tabular} & 
 \begin{tabular}[c]{@{}l@{}}UCI   (Breast \\      Cancer) \\      Non-temporal\end{tabular} \\ \hline
 
 \cite{zhou2020federatedcgai} & FCGAI & 2020 & 
 \begin{tabular}[c]{@{}l@{}}Air   quality monitoring \\      in non-i.i.d. conditions\end{tabular} & 
 \begin{tabular}[c]{@{}l@{}}Mask   recon. + Aux. data \\      (Multiple sources)\end{tabular} & 
 FCN, cWGAN & MAR & Air   quality \\ \hline
 
 \cite{li2020pbigan} & P-BiGAN & 2020 & 
 \begin{tabular}[c]{@{}l@{}}Continuous   MV from \\      irregularity in sampling\end{tabular} & 
 \begin{tabular}[c]{@{}l@{}}z-optimise   + Aux. model \\      (Classifier) + GAN invert\end{tabular} & 
 \begin{tabular}[c]{@{}l@{}}Continuous   CNN \\      layers, GAN\end{tabular},  & 
 \begin{tabular}[c]{@{}l@{}}Imbalanced   \\      (AUC)\end{tabular} & MNIST,   MIMIC-III \\ \hline
 
 \cite{liu2021glowimp} & GlowImp & 2021 & MVs   in MTS & z-optimise & 
 \begin{tabular}[c]{@{}l@{}}Invert.   1x1 CNNs, \\      norm. flows, WGAN\end{tabular} & 
 \begin{tabular}[c]{@{}l@{}}Imbalanced   \\      (AUC)\end{tabular} & 
 \begin{tabular}[c]{@{}l@{}}KDD   CUP 2018, \\      Challenge 2012 , \\      Health data\end{tabular} \\ \hline
 
 \cite{li2021eegcontextgan} & - & 2021 & 
 \begin{tabular}[c]{@{}l@{}}Incorrect   diagnosis \& \\      system failure in health \\      care applications\end{tabular} & 
 \begin{tabular}[c]{@{}l@{}}z-optimise   + Aux. \\      data \& model (Class.)\end{tabular} & 1-D   CNNs, WGAN & 
 \begin{tabular}[c]{@{}l@{}}Minority   class \\      oversampled\end{tabular} & Sleep \\ \hline

 \cite{feng2021cph} & CPH & 2021 & 
 \begin{tabular}[c]{@{}l@{}}Anaylsis   of population \\      Health data\end{tabular} & 
 \begin{tabular}[c]{@{}l@{}}Mask   recon. + Aux. \\      data (Class label)\end{tabular} & 
 \begin{tabular}[c]{@{}l@{}}2-D   CNNs, GAIN \\      (Feature x Time)\end{tabular} & - & 
 \begin{tabular}[c]{@{}l@{}}Ward   boundaries, \\      Chronic disease\end{tabular} \\ \hline
  
 \cite{awan2021imputationcgain} & cGAIN* & 2021 & 
 \begin{tabular}[c]{@{}l@{}}Class   imbalance, \\      class-specific \\      features for imputation\end{tabular} & 
 \begin{tabular}[c]{@{}l@{}}Mask recon. + Aux. \\      data (Classifier)\end{tabular} & 
FCN, GAIN& - & 5x   UCI, 4x Keel repo \\ \hline
 
 \cite{Kazemi2021igani} & IGANI & 2021 & 
 \begin{tabular}[c]{@{}l@{}}Reliability   of traffic \\      management\end{tabular} & GAN invert & 
FCN, WGAN & MCAR & Road   segments \\ \hline
  
 \cite{gao2022sgain_slim} & SGAIN* & 2022 & 
 \begin{tabular}[c]{@{}l@{}}Structural   health \\      monitoring\end{tabular} & 
 \begin{tabular}[c]{@{}l@{}}z-optimise  + Other \\      (Slim GAN)\end{tabular} & 
FCN, GAIN& 
 \begin{tabular}[c]{@{}l@{}}Continuous   \\      MVs\end{tabular} & 
 \begin{tabular}[c]{@{}l@{}}Bridge   (single \\      bridge)\end{tabular} \\ \hline
 
 \cite{xu2023mtgain} & 
 \begin{tabular}[c]{@{}l@{}}MT-\\      GAIN\end{tabular} & 2023 & Red   tide data analysis & 
 \begin{tabular}[c]{@{}l@{}}Mask   recon. + Aux. \\      model (Prediction)\end{tabular} & Conv/Deconv, GAIN & 
 \begin{tabular}[c]{@{}l@{}}Imbalanced   \\      (AUC), MCAR\end{tabular} & Red   tide \\ \hline
 
 \cite{huang2023tcgan2} & TCGAN* & 2023 & 
 \begin{tabular}[c]{@{}l@{}}General   purpose GAN \\      for improved supervised \\      tasks\end{tabular} & 
 \begin{tabular}[c]{@{}l@{}}Aux.   model (Clustering, \\      Classification)\end{tabular} & 1D-CNN, GAN & 
 \begin{tabular}[c]{@{}l@{}}Imbalanced   \\      (F1)\end{tabular} & 
 \begin{tabular}[c]{@{}l@{}}Synthesic   from \\      TimeGAN, \\      85x UCR time series\end{tabular} \\ \hline
 
 \cite{huang2023tsdigan} & 
 \begin{tabular}[c]{@{}l@{}}TS-\\      DIGAN\end{tabular} & 2023 & 
 \begin{tabular}[c]{@{}l@{}}Traffic   operations s.a. \\      congestion / speed \\      prediction, etc.\end{tabular} & 
 \begin{tabular}[c]{@{}l@{}}Data transform \\      (to image)\end{tabular} & Conv/Deconv, GAIN & - & 
 \begin{tabular}[c]{@{}l@{}}Performance   \\      Management \\      Systems\end{tabular} \\ \hline

 \cite{Bernardini2023ccgan} & ccGAN & 2023 & High   MVs in EHR & 
 \begin{tabular}[c]{@{}l@{}}Aux.   data (ITS with \\      least MVs)\end{tabular} & 
FCN, cGAIN& - & 
 \begin{tabular}[c]{@{}l@{}}Diabetic   (EHR), \\      MIMIC-III\end{tabular} \\ \hline
 
 \cite{Feng2023uaagain} & 
 \begin{tabular}[c]{@{}l@{}}UAA-\\      GAIN\end{tabular} & 2023 & 
 \begin{tabular}[c]{@{}l@{}}Pop.   health monitoring,\\       non-stationary \\      environment\end{tabular} & Uncertainty & 
FCN, GAIN& - & 
 \begin{tabular}[c]{@{}l@{}}Chronic   disease, \\      Hypertension, \\      Diabetes\end{tabular} \\ \hline
 
 \cite{Psychogyios2023igain} & I-GAIN & 2023 & 
 \begin{tabular}[c]{@{}l@{}}EHR   prediction \& \\      classification\end{tabular} & 
 \begin{tabular}[c]{@{}l@{}}z-optimise   + Aux. \\      model (Pre-imputation)\end{tabular} &FCN, GAIN& 
 \begin{tabular}[c]{@{}l@{}}BCE   on cat. data,\\       SMOTE, features \\      discarded\end{tabular} & 
 \begin{tabular}[c]{@{}l@{}}2x   Heart, Stroke, \\      PhysioNet Heart \\      failure\end{tabular} \\ \hline
 
 \cite{weng2024mviilgan} & MVIIL-GAN & 2024 & 
 \begin{tabular}[c]{@{}l@{}}Imbalanced,   incomplete \\      EHR, address problems \\      together for prediction.\end{tabular} & 
 \begin{tabular}[c]{@{}l@{}}z-optimise + \\      Mask recon.\end{tabular} & 
FCN, GAIN& 
 \begin{tabular}[c]{@{}l@{}}Minority   class \\      synthesised\end{tabular} & MIMIC-IV \\ \hline
 
 \cite{Hwang2024ccgain} & CC-GAIN & 2024 & 
 \begin{tabular}[c]{@{}l@{}}Energy   management in \\      smart grid environments\end{tabular} & 
 \begin{tabular}[c]{@{}l@{}}Mask   recon. + Aux. \\      model (Clustering,\\       classification)\end{tabular} &FCN, cGAIN & MCAR & Electricity \\ \hline
 
 \cite{ou2024rfgain} & RF-GAIN & 2024 & MV   from external factors & 
 \begin{tabular}[c]{@{}l@{}}Mask   recon. + Aux. \\      model (Pre-impute)\end{tabular} & 
FCN, GAIN& - & 
 \begin{tabular}[c]{@{}l@{}}Bearings,   \\      centrifugal pump\end{tabular} \\ \hline
 
 \cite{Li2024cgan_bayesian} & C-GAN & 2024 & 
 \begin{tabular}[c]{@{}l@{}}Quantify   uncertainty \\      with forecasting\end{tabular} & 
 \begin{tabular}[c]{@{}l@{}}Aux.   data (ext. data) \\      \& model (Likelihood \\      est.) + Uncertainty\end{tabular} & FCN, GAN & - & Air   pollution \\ \hline

\end{tabular}%
\footnotetext{\textbf{Abbreviations}: z-opimise \textit{- latent space optimisation approach;} Aux. \textit{- Auxiliary;} Rec. - \textit{recurrence-based approach;} Att. \textit{- Attention-based approach}\textit{;} recon. \textit{- reconstruction; }MV \textit{- missing values.}}
\end{sidewaystable}

\FloatBarrier

\section{Findings and future directions}  
\label{sec:discuss_future}
This paper reviewed GAN-based research on LDI from the past five years, including the most recent studies. Given our aims, we were able to 1) identify trends in approaches, methods and applications, 2) categorise the methods based on how they addressed the issue of missing values in longitudinal data, and 3) discover gaps and challenges, particularly regarding how GANs for LDI handle intrinsic features of longitudinal data like instance heterogeneity, and 4) examine how other data-level challenges like class imbalance and mixed data type that often co-exist with the missing data problem were managed. We present our findings here, along with potential future research directions.

\subsection{Meta-analysis of approaches, applications \& challenges in GANs for LDI}
GAN-based methods for LDI aimed to improve LDC and other downstream and analysis tasks varied primarily by 1) their GAN framework, training techniques and optimisation strategies employed, and 2) how missingness, whether caused by irregular sampling, drop outs or random occurrences, was addressed. We identified eight categories of approaches (as shown in Figure \ref{fig:meta_approaches}), 85\% of the works examined employ a combination of at least two approaches to address challenges associated with model training and handling sparse temporal data.

\subsubsection{Base neural architectures used in GAN for LDI}
\begin{figure}[!ht]
    \centering
    \includegraphics[width=0.8\linewidth]{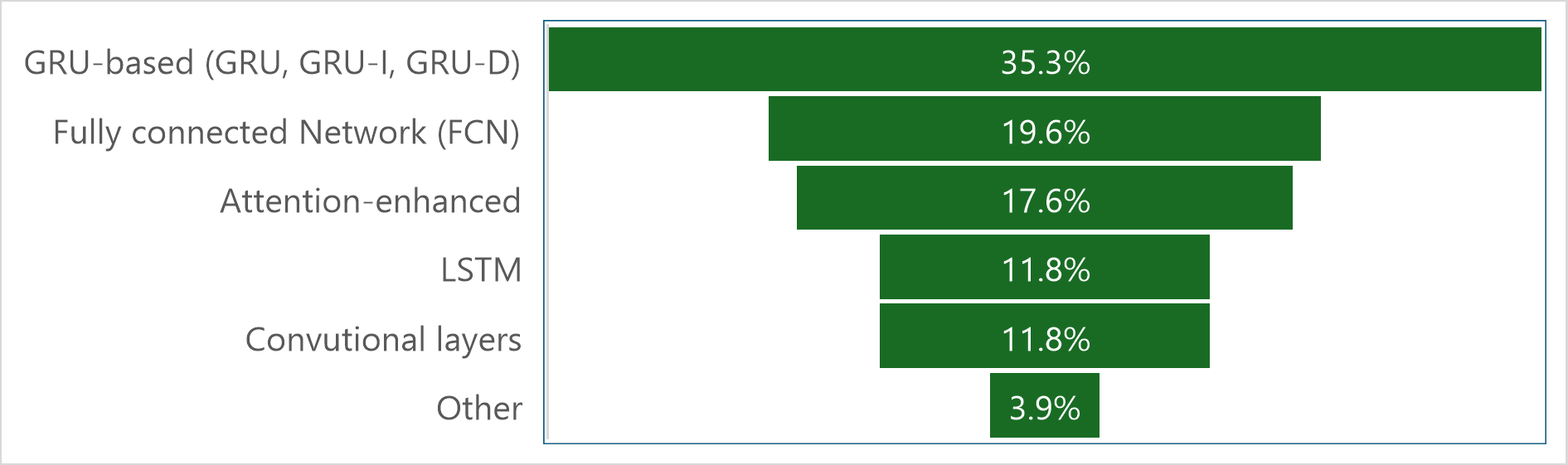}
    \caption{Base $G$ model of GANs for LDI}
    \label{fig:meta_base}
\end{figure}

Figure \ref{fig:meta_base} shows the distribution of base neural architectures used in the studies reviewed. Recurrence-based methods like RNNs are designed to capture temporal correlations within sequences but require equal-length sequences, making them suitable for regularly sampled data. 

To handle irregularly sampled time series, authors often discretised the data, introducing missing values. However, this sometimes result in highly sparse datasets, complicating the imputation process. Non-recurrence-based approaches, such as CNNs, can handle MTS of unequal lengths by processing them as individual univariate time series, sometimes referred to as (separate) channels. These are then combined through FCN layers to capture cross feature correlations. Not only is the non-i.i.d. assumption in longitudinal data violated here, but information on global temporal evolution lost. Nevertheless, FCN-based methods have shown strong performance in learning cross-feature dependencies, particularly in GAIN-based approaches.

Attention-based or attention-augmented methods offer an effective solution for handling irregular time series by learning both global and local correlations in the data. Despite progress, challenges persist, and there is a need for exploration of new frontiers that harness the potential of GANs and other deep generative models for LDI. Recent research has demonstrated the potential of large language models (LLMs) for time series data \cite{zhange2024llms_ts_survey} and pre-trained models for time series forecasting \cite{das2024foundationmodel_tsforecasting}. These new frontiers can be explored for transfer learning. Additionally, alternative architectures such as ODE-RNNs \cite{chen2018neuralode, rubanova2019odernn} and temporal CNNs \cite{lea2016tcn} could be further investigated for their applicability in GAN-based LDI.

\subsubsection{Evolution of methods over time}
\label{sec:meta_evolution_methods}
\begin{figure}
    \centering
    \includegraphics[width=1.0\linewidth]{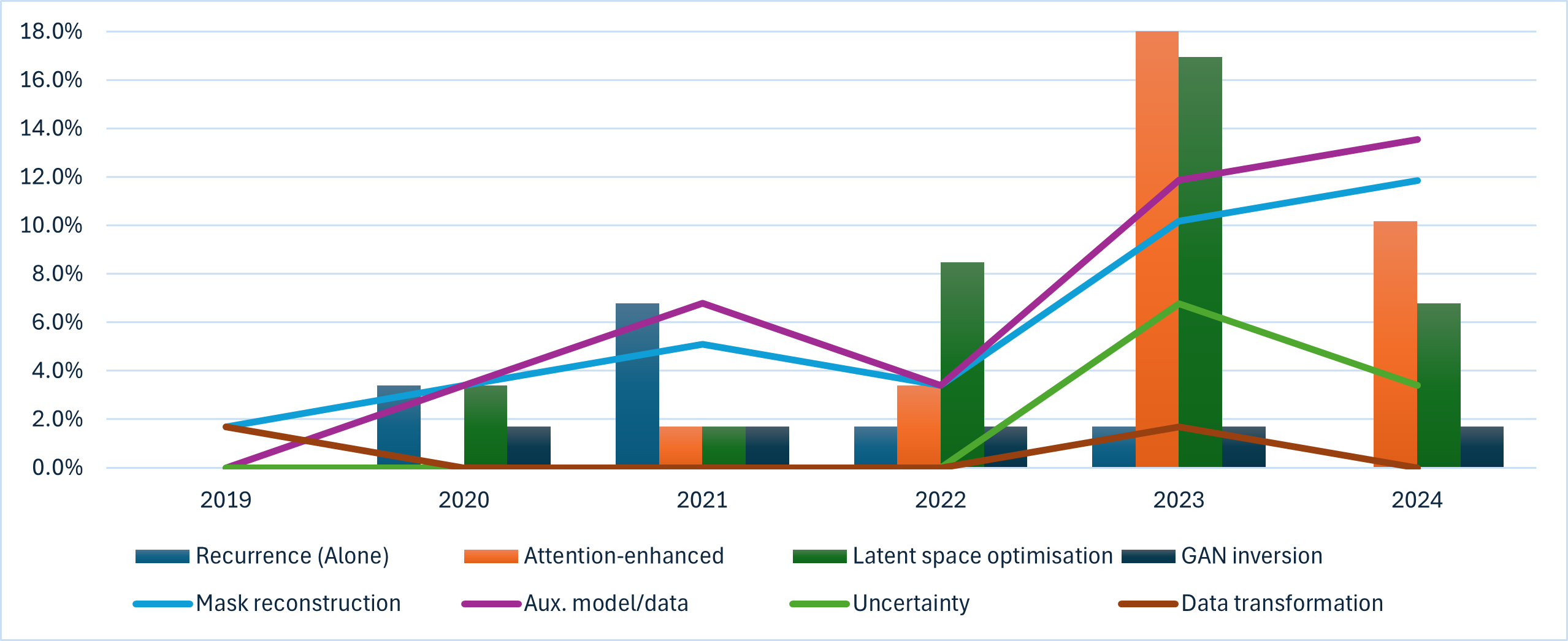}
    \caption{Approaches used in GANs for LDI. The Recurrence \enquote{Alone} category refers to methods that used RNNs like GRUIs in their conventional forms without other modifications or enhancements.}
    \label{fig:meta_approaches}
\end{figure}

Figure \ref{fig:meta_approaches} shows the evolution of GAN approaches for LDI over the years. Initially, there was a reliance on recurrence-based methods with next-step prediction and temporal decays - Recurrence (Alone) methods, but over time, there has been a shift towards more complex and hybrid methods. The adoption of attention mechanisms to better capture intricate temporal patterns in longitudinal data, and latent space optimisation for more robust sampling spaces have seen significant rises in application. Over time, focus has shifted towards GAIN-based mask reconstruction methods, and works leveraging auxiliary models and/or data to improve imputation. Interest in incorporating uncertainty into imputation and prediction for GAN-based LDI has only recently garnered interest. GAN inversion approach has remained consistently low, appearing in only 1\% of studies over the years, while novel techniques like data transformation remain scarce. Overall, 41\% of studies use auxiliary models/data, almost 40\% apply latent space optimisation, 37.5\% perform mask reconstruction, 10\% explored uncertainty, 9\% used GAN inversion, 4\% experimented with data transformation, and 5\% used other techniques like progressive growing GANs and federated learning for improved LDI in GANs.

\subsubsection{Application domains}
\begin{figure}[!ht]
    \centering
    \includegraphics[width=0.75\linewidth]{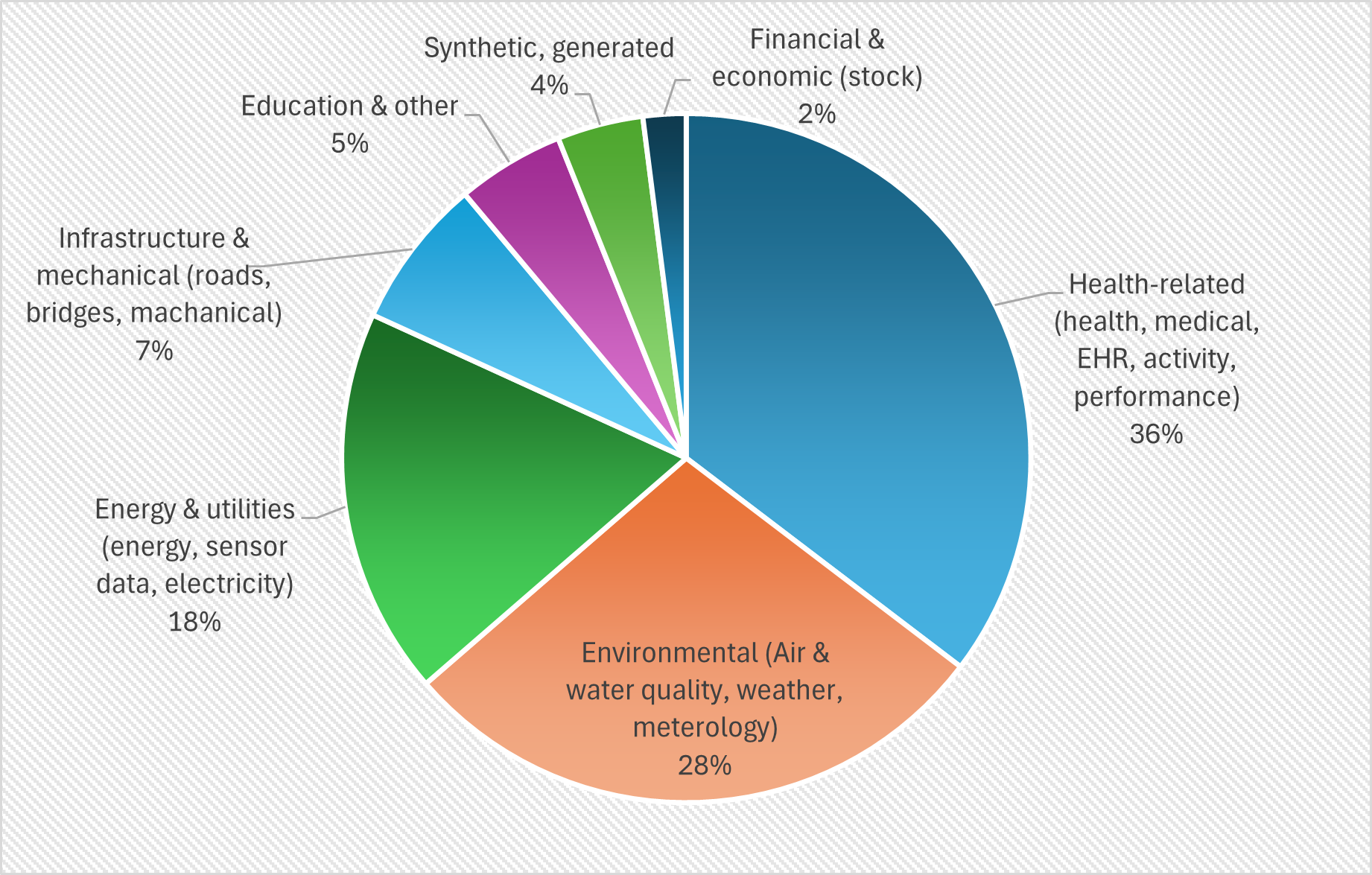}
    \caption{Application domains of studies using GANs for LDI}
    \label{fig:meta_data_applic}
\end{figure}

We focused on studies adhering to a definition of longitudinal data. For example, datasets like air quality across several cities were regarded as longitudinal since each city has unique characteristics like location and climate, leading to correlations between air quality features like humidity, CO$_2$ and temperature within individual cities. We also included studies on sensor emissions, mainly when data came from different localities, which was often the case. This allowed us to cover a wide variety of application domains. As Figure \ref{fig:meta_data_applic} shows, most GAN-based LDI methods have been applied in health and medical domains (36.5\%), followed by air quality and meteorological data, with fewer examples in Education and stock prices. 

\subsection{Key considerations for approach selection \& implementation}
\label{sec:strengths_challenges}
This section provides practical guidance on using GAN-based LDI approaches and an assessment of evaluation metrics for imputation. Each approach has strengths and limitations, and no single approach found to be universally optimal for all datasets or scenarios. Factors such as sequence length, missingness, irregular sampling, and ground truth availability significantly influence the choice of approach. Researchers often combine techniques to leverage complementary strengths. The selection should be guided by the dataset's characteristics and research objectives.

\subsubsection{Practical insights on approaches}
This section outlines critical insights for selecting and implementing the GAN-based LDI approaches described in Section \ref{sec:ldi_gan_approaches}.

\begin{enumerate}
    \item \textbf{Recurrence-based} methods are based on RNNs like LSTMs and GRUs, or their enhanced versions like GRU-I \cite{luo2018gruigan} and GRU-D \cite{che2017grud} for time series imputation. The generative capability of RNNs for seq2seq applications makes them a natural choice for time series imputation, thus applied in some 50\% of the studies. Key challenges, however, include vanishing gradients and bias explosion from accumulated imputation errors. Error propagation can be mitigated by next-step prediction \cite{cao2018brits}, with estimates validated by bi-directional RNNs \cite{Ni2022mbgan}, teacher-forcing \cite{Williams1989teacherforcing}, and temporal decay mechanisms \cite{che2017grud} that model the effects of time lags. Given their self-feeding nature, RNNs are prone to vanishing gradients, particularly with longer sequences. This can be alleviated with attention mechanisms \cite{oh2021sting} that focus on key time points and relevant feature dependencies, applying WGAN losses, and other solutions as proposed in Section \ref{sec:gan_train_challenges}. As RNNs take in equal time-length sequences, irregular time series would need to be discretised \cite{shukla2020surveyirregts}, with binary masks to indicate missing data (Eq. \ref{eqn:missmask}) and time lag information (Eq. \ref{eqn:timelag}). Where there exists an MNAR assumption, the masks would be used as model inputs. RNN-based models are often fitted with regression layers \cite{zhang2021e2ganrf} to improve next-step prediction and trained with additional reconstruction losses for imputation quality.
    \\
    \item \textbf{Attention-enhanced} methods often include attention mechanisms, their temporal variants, and transformer architectures \cite{Vaswani2017attention}. These methods outperform standard RNNs in capturing relevant intra-/inter-feature correlations and processing longer sequences through dynamic dependency learning. 15\% of works covered in the survey were attention-based, in particular on self-attention. Multi-head attention helps capture complex feature interactions \cite{ma2023ppcte-tsgain}, and variants like temporal attention target critical time steps, while sparse attention techniques \cite{zhou2021probsparse} can offer computational benefits for long sequences. In practice, attention layers, when combined with convolutional or recurrent architectures \cite{gao2024psacgan,sun2024dtin}, are able to learn long-range dependencies and inter-feature relationships. Extensions to the transformer architecture, such as with auto-regression \cite{bi2022geda}, can further enhance temporal learning. Despite their strengths, a key challenge exists in tuning hyperparameters, particularly the number of attention heads and sparsity thresholds that affect convergence and performance \cite{Xu2022ame}. Methods like Bayesian optimisation can be used to gain optimal model settings \cite{geng2024slimgain}, and strategies such as using data subsets \cite{Xie2020augment} and pre-trained components \cite{Liu2024transformerts} can help manage computational constraints.
    \\ 
    \item \textbf{Mask reconstruction}, inspired mostly by GAIN \cite{yoon2018gain}, assumes inherent data missingness and refines individual estimates for missing data elements iteratively during training. A key component that ensures the model learns the true data distribution is the hint mechanism, which should be empirically determined \cite{pingi2024faicgan}. This multiple imputation approach has increased in popularity by more than 50\% in the past year and a half (see Figure \ref{fig:meta_approaches}), and has been shown to be generally effective in informative missingness and high missing data rate conditions \cite{li2019misgan}. However, a major issue arises when the data's missing mechanism during training conflicts with GAIN's default MCAR assumption. While extensions of GAIN to incorporate MAR and MNAR assumptions with theoretical guarantees exist, 75\% of studies applying GAIN do not specify the assumed missing data mechanism. To improve robustness, works have incorporated auxiliary information, such as feature correlations or domain-specific patterns, to align the generator's learning process with the underlying missingness mechanism. Some 35\% of GAIN-based implementations follow the original framework with FCN layers. These tend to overlook temporal dependencies in the data but can be improved with the use of attention or recurrent layers.
    \\
    \item \textbf{Latent space optimisation} methods have gained momentum over the past two years due to their ability to create a structured latent space for reconstructing complex time series. These methods primarily employ AEs, or variants like DAEs \cite{Vincent2010dae} and VAEs \cite{kingma2013vae}, for their robust feature extraction and latent representation learning abilities. AEs can be jointly trained with the GAN \cite{luo2019e2gan}, or pre-trained \cite{wu2022imgan}, with latent encodings used by the AE decoder-based generator to generate imputation estimates. While pre-training simplifies processes, using complete samples to train the AE \cite{Li2024cgan_bayesian} assumes access to sufficient, complete, and representative data subsets. Transfer learning, where AEs are pre-trained on similar external datasets and fine-tuned on the target dataset, offers a viable alternative, but domain adaptation may prove necessary \cite{du21adarnn}. Controlling noise levels fed into the input data or latent space is necessary for learning robust representations \cite{vincent2008daechallenges}. Balancing multiple loss functions—adversarial, reconstruction, and latent space regularisation—is also critical for effective implementation. Weighted loss scaling \cite{Zadorozhnyy2021weightdisc} or multi-stage training, where components are initially trained separately before fine-tuning jointly, can stabilise training and improve convergence \cite{bi2022geda}. Incorporating temporal attention mechanisms into the AE can further enhance its ability to capture dependencies in sequential data, improving latent space optimisation for longitudinal datasets \cite{qin2023imputegan}.
    \\
    \item \textbf{Auxiliary data and/or model} approaches enhance imputation by leveraging external data or auxiliary modules, a methodology with a 65\% increase in adoption over the past two years. Auxiliary models, such as autoencoders or feature extractors, can be pre-trained on 1) complete subsets of data \cite{xu2023mtgain}, provided ground truth exists, or 2) external datasets for transfer learning \cite{Li2024cgan_bayesian}, though these must consider potential distribution shifts among the data sources \cite{du21adarnn}. To ensure alignment between the auxiliary model outputs and the GAN, domain adaptation techniques such as adversarial domain discriminators \cite{Nguyen2023ssdomadapt} are recommended. Or 3) pre-imputation estimates from methods such as K-NN or regression models \cite{ma2023ppcte-tsgain} to initialise imputation to stabilise training, but these should be validated against the dataset's missingness mechanism. Similarly, clustering techniques \cite{peng2024tscddgan} have been employed for contextual imputation. Static features \cite{pingi2024faicgan} and class labels \cite{Bernardini2023ccgan} can be used as auxiliary inputs to provide class-based imputation estimates.
    \\
    \item \textbf{Uncertainty-enhanced} methods quantify and leverage uncertainty in imputation estimates or prediction outcomes to improve these tasks. This approach has recently gained attention for its effectiveness in capturing temporal dependencies while providing uncertainty estimates across time points. Given GAN's multiple imputation capability, variability in imputation estimates, framed as model uncertainty, can be used for task optimisation \cite{Feng2023uaagain}. Balancing computational costs with the benefit of uncertainty quantification is important. An effective strategy is to use Monte Carlo (MC) dropout during inference to approximate uncertainty while limiting additional computational overhead \cite{Li2024cgan_bayesian}. Incorporating Bayesian layers into the generator can allow for uncertainty learning without requiring multiple imputations \cite{Nguyen2023lstmbnn}. To address the lack of standard uncertainty metrics, uncertainty can be evaluated using predictive entropy or mutual information metrics, ensuring robust validation \cite{Baboukani2020entropyts}.
    \\    
    \item \textbf{GAN inversion} refers to mapping generated samples back to the latent space to refine missing value estimates. Despite limited adoption (less than 10\% of studies), this approach offers a promising means to iteratively refine latent representations to improve imputation. Maintaining temporal dependencies between time points when mapping back to the latent space is a challenge, which is also a computationally expensive task. Here, image-based techniques, such as graph convolutions \cite{xu2023ganinvert}, pivot tuning \cite{roich2023pivotaltune}, and invertible generators \cite{Kazemi2021igani}, can be used. Lightweight architectures, such as recurrent layers or temporal graph networks, help reduce computational overhead while maintaining accuracy \cite{xu2023ganinvert}. Validation strategies should include temporal reconstruction metrics to ensure the preservation of time-series relationships \cite{wu2022imgan}.
    \\
    \item \textbf{Data transformation} reframes the problem of LDI into formats compatible with GAN-based frameworks. Two directions can be taken; in the first, ITS data can be transformed into image data \cite{huang2023tsdigan}, reframing LDI as an image inpainting task using imaging techniques such as Gramian Angular Fields \cite{Vargas2023gamts}. Preprocessing is essential to retain the temporal structure. Another direction involves aggregating or summarising temporal features, so that the problem is converted into a non-temporal tabular imputation task \cite{yang2019categoricalgain}. For tabular transformations, feature engineering is crucial to preserve temporal relevance, often through summarising time-series features into representative statistics \cite{Maximilian2018scale}. Post-transformation, attention mechanisms or graph-based encoders can be incorporated to reintroduce temporal dependencies, improving imputation quality \cite{xu2023ganinvert}.

\end{enumerate}

\subsubsection{Assessment of Evaluation Metrics}
This section provides an assessment of the evaluation metrics employed in the surveyed methods. These metrics are categorised into task-specific and domain-specific types. Task-specific metrics pertain to imputation, classification, and clustering tasks, as outlined in Tables \ref{tab:metric_impute}, \ref{tab:metric_class}, and \ref{tab:metric_cluster}. Domain-specific metrics, which are related to EEG signals and image datasets, are shown in Table \ref{tab:metric_domain}.

Metrics were categorised based on the prevalence of their use in the surveyed works \cite{Field2013metrics}. Metrics are considered \enquote{standard} if they are used by the majority of methods. Using established guidelines and statistical principles \cite{gliner2011research,muller2016introduction}, we use data thresholds align with prevalence of use and relevance across multiple studies.  

\begin{table}[!h]
\scriptsize 
\centering
\caption{Imputation Metrics by Domain}
\label{tab:metric_impute}
\renewcommand{\arraystretch}{1.3} 
\setlength{\tabcolsep}{6pt} 
\begin{tabular}{|p{1.2cm}|p{5.2cm}|p{3.4cm}|p{1.6cm}|}
\hline
\textbf{Abbrev.} & \textbf{Metric Name \& Description} & \textbf{Main Domains} & \textbf{Usage (\%)} \\ \hline
RMSE & \textit{Root Mean Squared Error}: Quantifies imputation performance, prioritizing larger errors. & Health-related, Traffic, Aeroengine, Energy, Air Quality & 80\% \\ \hline
MAE & \textit{Mean Absolute Error}: Measures average absolute errors for imputation accuracy. & Traffic, Energy, Health-related, Air Quality, Structural Health & 65\% \\ \hline
MSE & \textit{Mean Squared Error}: Evaluates squared differences between imputed and actual values. & Health-related (EHR, QAR), Traffic, Solar & 40\% \\ \hline
MRE & \textit{Mean Relative Error}: Assesses relative error to evaluate robustness. & Traffic, Air Quality & 15\% \\ \hline
MMD & \textit{Maximum Mean Discrepancy}: Measures distributional fidelity between real and imputed data. & Traffic, Sensor & 15\% \\ \hline
SMAPE & \textit{Symmetric Mean Absolute Percentage Error}: Evaluates percentage error under different missing rates. & Traffic, Health-related (Chronic Disease) & 10\% \\ \hline
R\textsuperscript{2} & \textit{R-Squared}: Measures the goodness of fit between imputed and original data. & Structural Health, Random Forest Models & 8\% \\ \hline
MNLL & \textit{Mean Negative Log-Likelihood}: Evaluates robustness, accounting for prediction uncertainty. & Structural Health & 5\% \\ \hline
\end{tabular}
\end{table}

In this study, metrics with more than 80\% usage, such as RMSE (80\%), are deemed universally accepted, or dominant practices. Metrics with usage rates between 50\% and 79\%, like MAE (65\%), can be classified as widely accepted but not universal. Metrics with usage 49-25\%\%, such as MSE (40\%) considered task-specific and those used in less than 25\% usage (like MRE (15\%)) be considered emerging or supplementary. This classification, as determined by metric usages aims to provide general guidance identifying trends in evaluation metrics used in time series and longitudinal data tasks.

\begin{table}[!h]
\scriptsize 
\centering
\caption{Classification Metrics by Domain}
\label{tab:metric_class}
\renewcommand{\arraystretch}{1.3} 
\setlength{\tabcolsep}{6pt} 
\begin{tabular}{|p{1.2cm}|p{5.6cm}|p{3cm}|p{1.6cm}|}
\hline
\textbf{Abbrev.} & \textbf{Metric Name \& Description} & \textbf{Main Domains} & \textbf{Usage (\%)} \\ \hline
AUC & \textit{Area Under the Curve}: Evaluates classification performance using ROC analysis. & \raggedright Healthcare (Mortality, Structured) & 50\% \\ \hline
Accuracy & \textit{Accuracy}: Assesses overall correctness of classification predictions. & \raggedright Healthcare, Traffic & 40\% \\ \hline
F1 Score & \textit{F1 Score}: Balances precision and recall for imbalanced datasets. & \raggedright Healthcare (Longitudinal) & 20\% \\ \hline
AUROC & \textit{Area Under Receiver Operating Characteristic}: Evaluates classification performance for imbalanced datasets. & \raggedright Healthcare (Mortality) & 15\% \\ \hline
PRAUC & \textit{Precision-Recall Area Under Curve}: Evaluates precision-recall trade-offs for imbalanced data. & \raggedright Healthcare & 5\% \\ \hline
MCC & \textit{Matthews Correlation Coefficient}: Measures the quality of binary classifications. & \raggedright Healthcare (Mortality) & 5\% \\ \hline
\end{tabular}
\end{table}

In most of the works covered, classification was performed as a downstream task to validate robustness of imputation methods in imbalanced datasets. AUC is the most widely adopted metric (50\%), particularly for imbalanced datasets. Accuracy (40\%), though common, is limited by its inability to account for imbalance. The usage trends of these and other metrics are shown in Table \ref{tab:metric_class}.

\begin{table}[!h]
\scriptsize 
\centering
\caption{Clustering Metrics by Domain}
\label{tab:metric_cluster}
\renewcommand{\arraystretch}{1.3} 
\setlength{\tabcolsep}{6pt} 
\begin{tabular}{|p{1.2cm}|p{5.6cm}|p{3cm}|p{1.6cm}|}
\hline
\textbf{Abbrev.} & \textbf{Metric Name \& Description} & \textbf{Main Domains} & \textbf{Usage (\%)} \\ \hline
CP & \textit{Clustering Purity}: Measures clustering consistency within true categories. & Sensor, Traffic & 5\% \\ \hline
NMI & \textit{Normalised Mutual Information}: Evaluates similarity between true \& predicted clusters. & Sensor, Traffic & 5\% \\ \hline
ARI & \textit{Adjusted Rand Index}: Adjusts for random clustering effects to enhance accuracy. & Sensor, Traffic & 5\% \\ \hline
RI & \textit{Rand Index}: Evaluates similarity between imputed \& original datasets. & Sensor & 5\% \\ \hline
\end{tabular}
\end{table}

For clustering and domain-specific metrics (Tables \ref{tab:metric_cluster} and \ref{tab:metric_domain}), the usage rates are too low to validate them as standards for evaluating imputation quality. However, they can still serve as supplementary metrics for integrated clustering tasks and domain-specific applications.

\begin{table}[!h]
\scriptsize 
\centering
\caption{Domain-specific Metrics by Domain}
\label{tab:metric_domain}
\renewcommand{\arraystretch}{1.3} 
\setlength{\tabcolsep}{6pt} 
\begin{tabular}{|p{1.2cm}|p{5.6cm}|p{3cm}|p{1.6cm}|}
\hline
\textbf{Abbrev.} & \textbf{Metric Name \& Description} & \textbf{Main Domains} & \textbf{Usage (\%)} \\ \hline
RUL Score & Root-Mean-Square Error for \textit{Remaining Useful Life}: Evaluates impact of imputed data for aeroengine lifespans. & Aeroengine Sensor & 10\% \\ \hline
FID & \textit{Fréchet Inception Distance}: Measures data quality \& diversity (e.g., EEG signals). & EEG Signals, Image & 5\% \\ \hline
IS & \textit{Inception Score}: Assesses quality \& diversity of imputed samples. & EEG Signals & 3\% \\ \hline
\end{tabular}
\end{table}

By introducing usage-based thresholds, this survey aims to provide a standardised framework for evaluating metrics across GAN-based imputation and classification methods, enabling clearer comparisons of performance under specific conditions.

\subsection{Main findings, gaps \& future directions}
This section presents our findings, highlights research gaps, and suggests future directions for LDC and existing LDI methods. Given our focus on addressing the missing data problem, we concentrated on reviewing GAN-based LDI methods. It was essential to consider 1) whether the missingness in the data stemmed from random processes, sampling irregularity or drop out, and 2) the assumed missing data mechanism, since these factors influence the imputation approach. We also examined how other data-level challenges like class imbalance and mixed data type were addressed, as like missing data, they can be considered intrinsic to longitudinal data. Of particular concern is the multi-dimensional nature of longitudinal data, and accounting for instance heterogeneity. 

\subsubsection{Missingness mechanisms \& assumptions}
Most GAN-based imputation models included in this paper do not specify the missing mechanism, if random masking is employed in the training regime, a MCAR is naturally assumed. While MCAR is the most benign and easiest mechanism to address, it is risky and often an unrealistic assumption in real-world longitudinal studies \cite{enders2011missing}. MCAR should still be imputed with consideration of complete cases \cite{demirtas2018flexibleimputation}, as done in some GAN-based LDI methods \cite{tang2020lgnet,zhou2020federatedcgai}. Challenges for LDI were seen to lie in 1) the missing data mechanisms or assumptions that ought define the imputation and evaluation strategies, 2) missing data patterns exhibited that may be informative to the modelling, 3) the assumed presence of ground truth, 4) and the data representation. A clear articulation of the missingness assumption under which imputation is performed is necessary. In our review, over 70\% of the studies did not specify a missing data assumption; those that did mostly assumed MCAR. However, as is common in most studies, randomly dropping observations for model training or evaluation enforces an MCAR assumption, which may not align with the missing data mechanism. Missing data patterns due to irregular sampling or dropout often carry important information. For instance, irregular sampling is described as \enquote{missing by design} \cite{groenwold2020missing} in longitudinal data like EHRs, while drop outs often follow a monotone missingness pattern \cite{zhang2020informdropout}, which some models \cite{pingi2024lecgan} attempt to consider by accounting for missing end-time points. Methods capable of handling long or continuous missing values \cite{Li2024bisrgain, wang2024wgainvae, qin2023imputegan} can be extended for dealing with missingness caused by drop outs. Models like GRU-D \cite{zhao2024saaegan} naturally assume MNAR. Tools like \cite{schouten2018generatemissing} that can enforce missingness patterns exhibited by MCAR, MAR and MNAR, and intentional masking under MNAR and MAR assumptions\cite{yoon2018gain} would be more effective than random masking applied during training or evaluation.  

Although GANs can marginalise out missing data and provide unbiased imputation estimates under the more relaxed MCAR assumption, this depends on the correct estimation of parameters by the GAN, which may not work well for temporal data with global dependencies. From a deep learning perspective, ensuring the MNAR assumption requires estimating imputes based on dependencies between global missingness pattern \cite{sun2020irregular} and global temporal correlations \cite{che2017grud,enders2011missing}. Models should also be evaluated based on missingness rates \cite{yoon2018gain}. Although it is important to note that multiple missing mechanisms can exist at varying degrees of missingness within a dataset \cite{Psychogyios2023igain}. Informative global missing data patterns between instances and groups \cite{little2013missing} as a research direction for LDI is also encouraged.

Although MAR is the common assumption in longitudinal datasets \cite{frees2004longitudinal}, MAR and MNAR are under-explored in GANs for LDI. Research shows that GAN models optimised for imputation may perform worse under the MNAR assumption than MAR or MCAR assumptions\cite{yoon2018gain}. This can create a misleading perception that better results under a particular missingness assumption means it exists for the data under study, and validates the imputation method and evaluation process. A more robust approach would be to assume MAR for longitudinal data \cite{demirtas2018flexibleimputation} and test the model's resilience under both MCAR and MNAR conditions. Also, a discrepancy exists between the missingness mechanism assumed by the model and that used during training or testing. For instance, applying random masking in GRU-D \cite{che2017grud}-based methods would cause the model to treat the random missing \enquote{patterns} as useful input. While some studies \cite{deng2022missmachs} propose theoretical extensions of the MCAR assumption, they do highlight the need to optimise GAN-based imputation under the MNAR assumption. Given the strict restrictions placed by MAR and MNAR on imputation methods, theoretical guarantees for the various LDI approaches are needed. 

In terms of the data representation mentioned in Definition 2 of Section \ref{sec:bg_ldc_defns}, different representations highlight various aspects of the data and pose unique challenges for GANs used in LDI \cite{shukla2020surveyirregts}. For example, the \textit{discretised} representation commonly used in RNNs can result in high levels of missing data and longer time lengths $T$ if there is significant variability in the sampling rates across variables. Using shorter time lengths $T$ to mitigate this issue often leads to more aggregation of values within intervals that can introduce imputation bias \cite{shukla2020surveyirregts}. GAIN-based models that apply this representation often include a reconstruction loss into the generator's objective to improve the accuracy of missing data estimates. On the other hand, CNN-based methods address irregular sampling by processing features with uneven time lengths as sets of (time, value) pairs. The challenge is that observations are temporally misaligned, and features are inter-dependent due to instance heterogeneity. While methods utilising attention mechanisms consider global correlations, this is a key research gap in other approaches that requires further exploration.

\subsubsection{Instance heterogeneity considerations}
Most existing methods, except \cite{pingi2024lecgan, pingi2024faicgan, Apalak2022sepsispred}, overlook static features and instance heterogeneity in the time series component. Accounting for instance-specific effects allows for variability and more precise estimates, leading to better inferences \cite{frees2004longitudinal}. While \cite{pingi2024lecgan, pingi2024faicgan} learned intra- and inter-relationships in static and temporal features through joint representation, it was not evident in \cite{Apalak2022sepsispred} how static features are integrated with the time series. Some time-series studies use static features solely for preprocessing \cite{ma2020maskehr} or treat temporal data similar to static data \cite{zhang2018stackelbergganmed} or visa versa. Given the flexibility of GANs, static features can condition the generated sample space to be jointly modelled with temporal features, as explored in \cite{yoon2019tsgan} or used to determine the inter-relationship of temporal features in the imputation task. Static variable stratification should be part of model evaluation \cite{bhanot2021staticeval,dash2020medgenerate} to improve longitudinal sample generation. Unfortunately, most works, even within health and medical domains, treat longitudinal datasets primarily as time series when addressing LDI, ignoring the effects of instance-specific features. 

Incorporating static features can provide more contextually meaningful imputation estimation, improve learning of variabilities for imputation estimates and enhance inference accuracy \cite{frees2004longitudinal}. Recent research \cite{pingi2024faicgan,Zhang2024pregating} conceptualises longitudinal data as multi-modal,  where one modality offsets the limitations of another in tasks like LDC. In contrast, other research \cite{AlQerem2023multidim} interprets multi-dimensionality as data having multiple features. Although we advocate for using static features to address instance heterogeneity, the related challenge of incorporating random effects into longitudinal data generation is yet to be explored in GANs. This can be achieved by treating static features as baseline characteristics of each instance or group. Clustering methods \cite{peng2024tscddgan,huang2023tcgan2}, federated GANs \cite{zhou2020federatedcgai}, and conditional GANs \cite{Hwang2024ccgain,Li2024cgan_bayesian} can be explored for modelling group effects. As a critical component of longitudinal data analysis, the incorporation of random effects in GANs with theoretical guarantees is a worthy research pursuit.

\subsubsection{Other data-level challenges}
The challenges of class imbalance and mixed data types, which often coexist with the missing data problem in longitudinal data were scarcely addressed by the GAN-based LDI methods reviewed. However, many studies deal separately with these issues, such as minority sample generation \cite{wang2023imbalancegan} and categorical embedding \cite{Yoon2023ehrsafe} for mixed data types. We explore here the effects of these challenges and suggest potential future research directions.

\textbf{Class imbalance }
poses a significant challenge for any deep learning-based method, including GANs, as they assume balanced distribution of data modalities \cite{chen2021imbalance}. Class-based profiling of instances could improve the quality of generated longitudinal samples. Only one study has attempted to jointly address the class imbalance and missing values \cite{awan2021imputationcgain}. Such end-to-end frameworks allow allow each task to leverage the other's learning \cite{kang2021rebootingacgan}; however careful regulation is needed in multi-task learning. A lack of representation from minority classes can result in insufficient latent encodings for augmentation. For LDI, a second-stage quality control measure for generated samples is recommended for GANs when augmenting minority samples to mitigate imbalanced LDC. 

Less than 15\% of studies reviewed used AUC to evaluate the robustness of their methods for minority class discrimination \cite{yang2023advrnnimpute,liu2021glowimp}. About 7\% of studies oversampled their data \cite{pingi2024faicgan,awan2021imputationcgain} using SMOTE \cite{Chawla2002smote}, which may have interpolated samples that lack diversity, leading to discriminator over-fitting, or generated samples may not be representative of the true distribution, particularly for high dimensional datasets \cite{Elreedy2024smotetheory}. Other considerations include determination of \enquote{most-suited} samples for imputation, based on instances' baseline features and class. Future research could also investigate causal representation learning \cite{Scholkopf2021causalrepresentlearn} to model static-temporal feature relationships. 

\textbf{Mixed data types }
exist in most longitudinal data with categorical or grouping information often in the static component. Proper representation of these non-numerical features is crucial for preserving feature correlations. Deep learning methods, including GANs, typically process continuous variables, as gradient descent requires computing gradients using continuous, differentiable functions. In GAN-based literature for LDI, there exits a gap in simultaneous generation (and imputation) of real-valued and non-real variables. Additionally, GANs face challenges in tabular data generation, including handling non-Gaussian, multi-modal and skewed distributions for continuous variables, as well as imbalanced discrete variables \cite{xu2019tabulargan}. These issues remain under-explored in GAN-based LDI methods. 

Among the GAN models reviewed, most did not specify their approach to treating categorical features. In works covered, \cite{yang2019categoricalgain} converts categorical features into fuzzy representations, \cite{Psychogyios2023igain} applies binary cross entropy for categorical data, \cite{Xiang2020ehrembed} does 2-D embedding of categorical features to maintain their hierarchical structures, and \cite{li2023ehrgenmixeddata} uses pre-trained dual VAEs for continuous and discrete data. While advances exist in this area, how feature correlations and imputation are affected by these strategies need to be analysed further.

\subsubsection{Other considerations}
It is important to highlight additional considerations that can significantly impact the success in future development of GAN-based LDI methods and warrant further exploration. 

\textbf{Lack of ground truth } 
can be defined as the insufficient number of complete records in a dataset to provide a representative subset of the hypothetically complete dataset. Relying on ground truth is considered unrealistic for longitudinal datasets \cite{enders2011missing} as it leads to biased parameter estimation \cite{dong2013missing}. Some methods split their data into complete and incomplete subsets, using the complete set for model training, while other studies omitted incomplete cases altogether. Both approaches assume the presence of ground truth in longitudinal data, limiting the generalisability of their designs and findings to most longitudinal data, like health data. The rising popularity of GAIN-based methods attests to the widespread issue of inherent data missingness that need further investigation of methods. 

\textbf{Capturing temporal transitions }
and maintaining accurate transitions within time series data\cite{Jarrett2021tsgenconstrastive} is one of GANs' shortcomings in generating viable estimates for LDI. Discriminators help generate globally coherent time series, however GAN's implicit temporal data learning makes it difficult to verify the accuracy of individual transitions. This lack of explicit step-wise transition can lead to error propagation in longer sequences, failing to model the real data's overall behaviour. Techniques like professor forcing \cite{cao2018brits}  mitigates this, however investigations are need to ensure the joint distribution of multiple-step prediction in LDI GANs \cite{zhang2021e2ganrf} or performing transitionary validation of the entire sequence \cite{Coletta2023constrainedtsgen, Jarrett2021tsgenconstrastive}.

\textbf{Standard measures for quality } 
of generated longitudinal data is necessary. The \enquote{quality} of imputation estimates is often judged by the GAN's ability to reconstruct observed data. Apart from this quantitative analysis based on RMSE and that of predictive accuracy measures (like ROC-AUC and F1 scores), other methods developed of assessing the quality of imputed samples include 1) interpolation of latent points to demonstrate smooth variations in the sample space that show a continuous and meaningful latent space \cite{guo2019mtsgan}, 2) congeniality tests that ensure imputed values respect feature-label relationships \cite{yoon2018gain}, and 3) comparing similarity between real and learned distributions using metrics like W-loss \cite{arjovsky2017wgan}. However, there is no universally accepted set of metrics for quality of imputed estimates, that is, their \enquote{best-fittedness} for imputation; further research is needed to fill this void.

\textbf{Joint optimisation }
of auxiliary tasks like classification, clustering, or minority class generation can be challenging in GANs for LDI, although shown to improve primary objectives like imputation or augmentation \cite{kang2021rebootingacgan,ramsundar2015multitasknetworks}. However, the learned distribution should most accurately reflect the real distribution, otherwise the resulting model will be biased \cite{shu2017acganbias}. Of the methods surveyed, only 35\% incorporated an auxiliary model, typically for prediction or classification. The challenge lies in the increased complexity and additional hyper-parameters for fine tuning, leading to higher optimisation costs. Carefully designed sanity checks and ablation studies are essential for a thorough assessment of such models. Most works that applied joint training objectives related to classification, clustering, and forecasting. Auxiliary classification tasks helped mitigate propagation of imputation errors, allowing class-based imputation and early classification. These benefits present valuable opportunities for further exploration, particularly with the use of uncertainty measures.

\section{Conclusions} \label{sect:future}
This paper provides a comprehensive review of how GANs have been applied to address the data-level challenges in LDC, focusing on handling missing values due to random processes, irregular sampling and study drop outs. Recognizing that mixed data types are a challenge for GANs and class imbalance is a confounding issue in LDC, often exacerbated by missing values; we closely examine how these challenges were addressed in the reviewed studies. Through an in-depth analysis of state-of-art methods, we propose eight categories for GAN-based LDI methods and compare how methods within each approach address data-level challenges posed by LDC. The most common approaches involve the use of auto-encoders for more structured latent representation for reconstruction of sparse data, and mask reconstruction for handling inherent data missingness. These are followed by methods applying 1) attention mechanisms for learning long sequences and sparse representations, 2) auxiliary models for feature selection, pre-imputation estimates or classification/clustering, and 3) auxiliary data, particularly for class-based imputation and transfer learning. Approaches relying on data transformations, GAN inversion and uncertainty estimations are less common. We also highlight research gaps and future research directions.

We acknowledge that considerations for class imbalance, mixed data types and integration of static features in GAN-based LDI can be complex and resource-intensive. Each approach and training strategy offers benefits and has drawbacks. The choice between joint training or a two-phased imputation strategy, enhancements for boost performance for temporal learning, or relieve training complexity all come with training challenges and  resource and complexity considerations that can be difficult to measure against GAN's inherent instability, as discussed in Section \ref{sec:gan_train_challenges}. Complex models often need finely tuned hyper-parameters, efficient training and appropriate evaluation strategies. Efforts in model simplification can solve these issues to some extent \cite{Vaswani2017attention}, but strategies like transfer learning or use of pre-trained auxiliary models have to be weighed against risks for information loss in cases where missing data patterns are informative \cite{little2019missing} or where distribution shifts may arise \cite{du21adarnn}.

As a byproduct of this analysis, the paper reveals the versatility, extensibility and potential of GANs to handle intricacies of missing longitudinal data analysis, in order to improve it's quality and usability for downstream tasks such as classification. The general lack of consideration for key longitudinal data properties, such as multi-dimensionality, instance heterogeneity, class imbalance, missing data assumption, inter-feature correlations, and the generation of mixed data types in GAN-based LDI methods likely stems from the complexity of jointly addressing these issues, despite extensive literature on GANs addressing these problems individually. Overall research gaps for GAN-based imputation of longitudinal data includes effective use of static variables to improve task objectives, mixed data type imputation, accounting for global temporal and missing data dependencies, providing theoretical guarantees for imputation under MNAR and MAR conditions, need for effective and standardised evaluation strategies for quality of reconstructed or imputed samples (beyond standard reconstruction based losses), and analysis of \enquote{best sample matches} for imputation. We hope the discussions and knowledge gaps highlighted in this paper will guide future research and advancements in the field.

\clearpage


\bibliographystyle{plain}
\bibliography{sn-bibliography}

\end{document}